\documentclass[10pt,journal,compsoc]{IEEEtran}
\ifCLASSOPTIONcompsoc
  \usepackage[nocompress]{cite}
\else
  \usepackage{cite}
\fi

\ifCLASSINFOpdf
\else
\fi

\usepackage[utf8]{inputenc}
\usepackage{xcolor}
\newtheorem{definition}{Definition}
\usepackage{amsmath}
\usepackage{graphicx}
\usepackage{url}

\usepackage{float}
\usepackage{amssymb}
\usepackage{tabularx}
\newcolumntype{L}{>{\raggedright\arraybackslash}X}
\usepackage{textcomp}
\usepackage{subfig}
\usepackage[font=scriptsize]{caption} 
\usepackage{enumitem}
\definecolor{PineGreen}{rgb}{0.0, 0.47, 0.44}

\begin{document}
%
\title{Deep Learning for Predictive Business Process Monitoring: Review and Benchmark}
%
%
%
%

\author{Efrén~Rama-Maneiro,
        Juan~C.~Vidal,
        and~Manuel~Lama
        \IEEEcompsocitemizethanks{
          \IEEEcompsocthanksitem E. Rama-Maneiro is with the Centro Singular de Investigación en Tecnoloxías Intelixentes (CiTIUS), Universidade de Santiago de Compostela, 15782, Santiago de Compostela, SPAIN. email: efren.rama.maneiro@usc.es \protect \\
          \IEEEcompsocthanksitem J. C. Vidal is with the Centro Singular de Investigación en Tecnoloxías Intelixentes (CiTIUS) and with Depeartamento de Electrónica e Computación, Universidade de Santiago de Compostela, 15782, Santiago de Compostela, SPAIN email: juan.vidal@usc.es \protect \\
          \IEEEcompsocthanksitem M. Lama is with the Centro Singular de Investigación en Tecnoloxías Intelixentes (CiTIUS), Universidade de Santiago de Compostela, 15782, Santiago de Compostela, SPAIN. manuel.lama@usc.es \protect \\
        }
\thanks{Manuscript received \today~revised X}}


\IEEEtitleabstractindextext{%
\begin{abstract}
   Predictive monitoring of business processes is concerned with the prediction of ongoing cases on a business process. Lately, the popularity of deep learning techniques has propitiated an ever-growing set of approaches focused on predictive monitoring based on these techniques. However, the high disparity of process logs and experimental setups used to evaluate these approaches makes it especially difficult to make a fair comparison. Furthermore, it also difficults the selection of the most suitable approach to solve a specific problem. In this paper, we provide both a systematic literature review of approaches that use deep learning to tackle the predictive monitoring tasks. In addition, we performed an exhaustive experimental evaluation of 10 different approaches over 12 publicly available process logs.
\end{abstract}

}

\maketitle

\IEEEdisplaynontitleabstractindextext

%
\IEEEpeerreviewmaketitle

\section{Introduction}

Process mining is a discipline concerned with analyzing process logs to either discover representations of the underlying process model (\textit{process discovery}), to compare the process model with an event log of the same process (\textit{process conformance}), or to improve the business process (\textit{process enhancement})~\cite{Aalst2011}. The starting point of process mining techniques is the event log, which is a set of events with at least a \textit{case identifier}, an \textit{activity}, and a \textit{timestamp}, and, optionally, \textit{case attributes}, which are shared by all the events of the same case, or \textit{event attributes}, that are specific of each event. \tablename~\ref{tab:log} shows an example log from the set of procedures applied to patients in a hospital. This event log provides information about each case identifier, activity, timestamp and resource of each event, i.e, who executed each activity.

\begin{table}[htb]
  \scriptsize
\caption{Excerpt of a business process log}~\label{tab:log}
\begin{tabular}{l|l|l|l}
  \hline1
Case ID  & Activity              & Timestamp           & Resource    \\ \hline
Case2118 & Inclusion & 14-01-2010 07:52:50 & Peter \\
Case2118 & TAC & 09-02-2010 13:01:11 & Joseph \\
Case2118 & Blood Analysis & 17-02-2010 07:44:53 & Joseph  \\
Case2118 & Intervention & 17-02-2020 07:44:59 & Joseph \\
Case2118 & Discharge patient & 18-02-2020 09:00:10& Joseph \\ \hline
Case2088 & Inclusion & 04-02-2010 08:37:45 & Peter  \\
Case2088 & Pneumothorax scan & 04-02-2010 09:01:28 & Peter  \\
Case2088 & TAC & 04-02-2010 09:01:35 & Peter  \\
Case2088 & Blood Analysis& 16-03-2010 13:08:40 & Peter  \\
Case2088 & Discharge patient & 31-03-2010 11:08:53 & Dio  \\ \hline
\end{tabular}
\end{table}

Predictive monitoring is a subfield of process mining that aims to forecast relevant information of an ongoing business process case~\cite{Maggi2014}. Taking as an example the log of \tablename~\ref{tab:log}, we might want to forecast which would be the next medical trial applied to a patient, the remaining time until a patient is discharged, or whether a patient is going to be intervened. These forecasting is performed by means of a classifier trained on the ongoing running cases (also called \textit{prefixes}). A wide array of traditional machine learning techniques has been applied to predictive monitoring such as Transition Systems~\cite{Aalst2011a, Senderovich2015}, Probabilistic Finite Automatons~\cite{Becker2014, Breuker2016}, or Support Vector Machines~\cite{Kang2012, Cabanillas2014}.

More recently, deep learning and neural networks have gained a lot of attention due to their success in fields such as computer vision~\cite{He2016, Simonyan2015} or natural language processing (NLP)~\cite{Vaswani2017, Sutskever2014}. Due to the sequential nature of business processes, recurrent neural networks (RNN) were found to be a good fit to approach predictive monitoring~\cite{Evermann2017, Francescomarino2017}. Nowadays, many predictive monitoring approaches are based on deep learning due to their promising results against more classical machine learning techniques~\cite{Tax2018}. Unfortunately, the high number of neural network architectures, ways to encode the partial traces and events, the number of predictive tasks available, and, sometimes, the difficulty of quantifying the differences and contributions between the works may complicate the task of defining what has already been done, what can be researched and future research directions. Furthermore, and more importantly, approaches are applied to a reduced number of datasets and with significantly different experimental setups. This makes it difficult to new researchers that want to compare their approach with a strong baseline of state-of-the-art works to both the necessity of adapting the experimental setups to be fair and the high computational resources that characterize the usage of deep learning approaches.

    Thus, the need for a review is much-needed to tackle these issues. In this paper, we identify and provide a taxonomy for the most relevant deep learning approaches for predictive monitoring. Then, using the set of identified approaches as a starting point, we select those that have their implementation available and perform an experimentation and a statistical comparison in a fair setting using a wide number of process logs. Then, we use the results to highlight the most relevant differences between them. We have made public the source code\footnote{The code link is available in the supplementary material.}, trained models and results of the experimentation in the hope that it serves as a starting point for new researchers in predictive monitoring.

The rest of the paper is structured as follows: Section 2 formally defines the most common predictive monitoring problems. Section 3 highlights the difference between this survey and other predictive monitoring surveys. Section 4 highlights the search methodology to perform the systematic literature review. Section 5 shows the classification and taxonomy of the retrieved studies, explaining its main achievements. Section 6 shows the experimental setup, results, and discussion of the benchmarking of the available approaches. Finally, Section 7 concludes the paper highlighting future lines of work.

\section{Definitions}
\label{sec:background}
Given a certain event prefix of a running case, predictive monitoring is concerned with forecasting how different aspects of the next event or sequence of events will unfold until the end of the case. Formally, $A_n, D^{i}_{n}, T_n, O_n$ be the activity, attribute $i$, timestamp, and outcome of the event $e_n$. Let $hd^k(\sigma)$ be a event prefix such as $hd^k(\sigma) = \langle e_1, \ldots, e_k \rangle$, $e'$ be a predicted event by a function $\Omega$, and let $\oplus$ be the concatenation operator between two sequences, then, depending on the predictive task at hand, we can define the following functions $\Omega$:

\begin{definition}
  The next activity prediction problem can be defined as $\Omega_A(hd^k(\sigma)) = A'_{k+1}$.
\end{definition}
\begin{definition}
  The next attribute, $d_i$, prediction problem can be defined as $\Omega_{D^{i}}(hd^k(\sigma)) = (D')^{i}_{k+1}$.
\end{definition}
\begin{definition}
  The next timestamp prediction problem can be defined as $\Omega_T(hd^k(\sigma)) = T'_{k+1} - T'_{k}$.
\end{definition}

\begin{definition}
  The outcome of an event prefix can be predicted as $\Omega_{O}(hd^k(\sigma)) = O'_{k} $.
\end{definition}
The activity suffix prediction problem can be defined in two different ways: as the recursive prediction of $A$, i.e, using the newly predicted activities as new inputs for the next prediction until the end of the case (``[EOC]'') is reached (Definition~\ref{def:sa-recursive}) or directly predicting the activity suffix (Definition~\ref{def:sa-direct}).

\begin{definition}
  \label{def:sa-recursive}
   An activity suffix can be recursively predicted as $\Omega_{SA} = \langle \Omega_{A}(\sigma ') = A'_i  \: |  \:  \sigma ' = hd^k(\sigma) \oplus \langle e'_{k+1}, \ldots, e'_{i-1} \rangle \rangle$ while $A'_i \neq [EOC]$.
  \end{definition}
  \begin{definition}
    \label{def:sa-direct}
    An activity suffix can be directly predicted $\Omega_{SA}(hd^k(\sigma)) = A'_i \ldots A'_k$.
  \end{definition}

Functions for predicting an attribute suffix and remaining time can be defined analogously.

\begin{definition}
  An attribute suffix can be recursively predicted as $\Omega_{SD_{j}} = \langle \Omega_{D_{j}}(\sigma') = D^{j}_{i} \: |  \: \sigma' = hd^k(\sigma) \oplus \langle e'_{k+1}, \ldots, e'_{i-1} \rangle \rangle$ while $A'_i \neq [EOC]$.
\end{definition}
\begin{definition}
  An attribute suffix can be directly predicted as $\Omega_{SD_{j}}(hd^k(\sigma)) = D'^{j}_{i} \ldots D'^{j}_{k}$.
\end{definition}

\begin{definition}
  \label{def:remaining-time}
  Let $\theta$ be the sequence of predicted next timestamps such as $\theta = \langle \Omega_T(\sigma ') = T'_i - T_{i-1} \: | \: \sigma ' = hd^k(\sigma) \oplus \langle e'_{k+1}, \ldots, e'_{i-1} \rangle \rangle $ while $ A'_i \neq [EOC] $, then the remaining time can be calculated as $\Omega_{RT}(hd^k(\sigma)) = \sum_{i=k}^{n} \theta_i$.
\end{definition}

\begin{definition}
   \label{def:remaining-time-direct}
The remaining time can be directly calculated as $\Omega_{RT}(hd^k(\sigma)) = T'_{n} - T_{k}$ where $T'_n$ denotes the predicted timestamp for the last event.
\end{definition}

\section{Related work}
Several authors have addressed the problem of reviewing the current state of the art in predictive monitoring. In~\cite{MarquezChamorro2018}, the authors review a set of studies and classify them in process-aware methods and non-process aware methods, depending on whether they require a process model as an input or not, and also whether they treat predictive monitoring as a regression problem or as a classification problem. In~\cite{Francescomarino2018}, the authors perform a systematic literature review with the aim of helping companies to select their best suited predictive monitoring framework according to multiple dimensions such as the predictions made by the approach, the domain where it is applied, the algorithm developed, and the input used. In~\cite{Harane2020}, the authors perform a study of three deep learning approaches for predictive monitoring, taking into account different dimensions, such as the encoding scheme or the prediction target.

Regarding to benchmarking studies, in~\cite{Tama2019}, the authors compare the performance of 20 different supervised learning classification techniques over 6 different process logs to predict the next event in a business process. They conclude that the best machine learning classifier, in terms of accuracy, is the credal decision tree.

In~\cite{Tax2018}, a comparison of the following families of techniques for the next activity prediction problem is made: \textit{(i)} recurrent neural networks, \textit{(ii)} markov models, \textit{(iii)} grammar induction techniques, which learn a set of production rules that describe a language (in this case, a process log), \textit{(iv)} process discovery-based techniques, and \textit{(v)} automata based prediction techniques.

In~\cite{Verenich2019}, the authors focus their study on the prediction of the remaining time
in a business process, comparing several bucketing, encoding, and
supervised learning techniques in terms of the Mean Absolute Error of the
predictions. They also compare the supervised learning techniques with 3 different process-aware methods. Their conclusion is that LSTMs outperform other approaches for remaining time prediction. In~\cite{Teinemaa2019}, the authors provide a systematic literature review of outcome-oriented predictive monitoring and compare the performance of four different machine learning techniques (random forest, logistic regression, support vector machines and extreme gradient boosting) with multiple encodings to predict the outcome of a business process. In~\cite{Kratsch2020}, the authors perform a similar analysis for outcome prediction but comparing another set of machine learning techniques (random forest, support vector machines, deep feedforward networks, and LSTMs).

  In~\cite{Neu2021} they perform a similar review to this one, but they do not provide any fair benchmarking of the approaches. In~\cite{Weytjens2020} they perform an experimentation to compare CNN, LSTM, and LSTM with an attention mechanism to predict the outcome of a running case. They conclude that CNN perform similar to LSTMs, and, since the former is computationally faster, it should be the architecture to use. In~\cite{Heinrich2021} they introduce key-value-predict attention networks to improve the performance of LSTMs, and introduce the usage of process data properties as additional features. However, their experimentation is limited to a small number of approaches and not every approach is tested in every proposed dataset.

  Only~\cite{Harane2020} and~\cite{Neu2021} have similar objectives as the review conducted in this paper. In~\cite{Harane2020}, the analysis is limited to the first three deep learning approaches in predictive monitoring, but without any benchmarking, so the results reported are taken directly from the papers. In~\cite{Neu2021}, even though a more thoroughly review is provided, the analysis also lacks any benchmarking between the approaches.

  Other authors include Deep Learning approaches when comparing their solution with the state-of-the-art in predictive monitoring. However, they limit their comparison to evaluating LSTM~\cite{Tax2018, Verenich2019, Kratsch2020}, and GRU~\cite{Tax2018} alongside other machine learning algorithms. No other review study benchmarks the original deep learning implementations of the state-of-the-art approaches under controlled conditions to provide a fair comparison between them.

\section{Analysis and classification}
We performed an analysis and classification of the predictive monitoring that use deep learning\footnote{The details of the search methodology are available in the supplementary material.}. The results of this classification are available in \tablename~\ref{tab:studies}. \tablename~\ref{tab:morphological} reflects every possibility for each of the classification criteria.

\begin{itemize}
  \item \textit{Input data}: data from the business process logs used to train the predictive model.
  \item \textit{Predictions}: the elements of the events that the neural network is trained to forecast.
  \item \textit{Neural Network Type}: type of neural network used such as feedforward, autoencoder, convolutional, recurrent or transformer.
  \item \textit{Sequence encoding}: how event prefixes are converted into
    learnable tensors by the neural net.
  \item \textit{Event encoding}: how each individual categorical and continuous variable is encoded in a feature vector.
\end{itemize}

\begin{table*}[!th]
  \scriptsize
  \centering
  \begin{tabular}{lcp{1.4cm}p{1.3cm}p{1.3cm}p{1.5cm}lll}
    \hline
    Author, Year & Reference & Network type & Sequence encoding & Event encoding & Input data & Time features &  Prediction \\ \hline

    Evermann et al., 2017 & \cite{Evermann2017} & LSTM & CONT & EMB & ACT & - & ACT \\

    Francescomarino et al., 2017 & \cite{Francescomarino2017} & LSTM & PRFX & OH & ACT, LTL & TSP, TSSC, WK, TMD & SFX \\

    Mehdiyev et al., 2017 & \cite{Mehdiyev2017} & AE + DFNN & NGRAM & - & ACT, ATTR & - & ACT \\

    Tax et al., 2017 & \cite{Tax2017} & LSTM & PRFX & OH & ACT & TSP, TSSC, WK, TMD & ACT, NT, SFX, RT \\

    Navarin et al., 2017 & \cite{Navarin2017} & LSTM & PRFX & OH & ACT, ATTR & TSP, TSSC, WK, TMD & RT \\

    Al-Jebrni et al., 2018 & \cite{Al-Jebrni2018} & CNN & CONT & EMB & ACT & - & ACT \\ 

    Khan et al., 2018 & \cite{Khan2018} & DNC & PRFX & OH & ACT & TSP, TSSC, WK, TMD & ACT, NT, RT, SFX \\

    Metzger et al., 2018 & \cite{Metzger2018} & LSTM & PRFX & OH & ACT & - & OUT\\

    Schönig et al. 2018 & \cite{Schoenig2018} & LSTM & CONT & OH & ACT, ATTR & - & ACT, RES\\ 

    Mehdiyev et al., 2018 & \cite{Mehdiyev2018} & AE + LR & NGRAM & - & ACT, ATTR & - & ACT \\

    Camargo et al., 2019 & \cite{Camargo2019} & LSTM & PRFX & P-EMB & ACT, R & TSP & ACT, ROLE, NT, RT, SFX, RLSFX \\

    Lin et al., 2019 & \cite{Lin2019} & LSTM & CONT & EMB & ACT, ATTR & - & ACT, ATTR, SFX, ATTRSFX \\

    Hinkka et al., 2019 & \cite{Hinkka2019} & GRU & PRFX & OH & ACT, ATTR, C-ATTR & - & ACT [CP] \\

    Theis et al., 2019 & \cite{Theis2019} & DFNN & TSS & FB & ACT, PM, ATTR & DTPA & ACT \\

    Wahid et al., 2019 & \cite{Wahid2019} & DFNN & SE & EMB & ACT, ATTR & TSP, TSSC & RT \\

    Pasquadibisceglie et al., 2019 & \cite{Pasquadibisceglie2019} & CNN & PRFX & FB & ACT & TSSC & ACT \\

    Wang et al., 2019 & \cite{Wang2019} & B-LSTM & PRFX & OH & ACT, ATTR & - & OUT \\ 

    Mauro et al., 2019 & \cite{Mauro2019} & CNN & PRFX & EMB & ACT & TSP & ACT \\

    Folino et al., 2019 & \cite{Folino2019} & LSTM  & PRFX & OH, EMB & ACT, ATTR & - & OUT \\

    Hinkka et al., 2019 & \cite{Hinkka2019a} & GRU/LSTM & CONT & OH & ACT & - & OUT\\ 

    Philipp et al., 2020 & \cite{Philipp2020} & TRANS & CONT & EMB & ACT & - & ACT \\ \hline

  \end{tabular}
  \caption{Collected studies for Predictive Monitoring that use Deep Learning.}~\label{tab:studies}
\end{table*}

\begin{table*}[!th]
  \centering
  \scriptsize
  \begin{tabular}{|p{2.2cm}|p{2.2cm}|p{2.2cm}|p{2.2cm}|p{2.2cm}|p{2.2cm}|}
    \hline
    Network type & Sequence encoding & Event Encoding & Input data & Time features & Prediction \\ \hline
    Long Short Term Memory (LSTM) & Continuous (CONT) & Embedding (EMB) & Activity (ACT) & Time since start of the case (TSSC) & Activity (ACT) \\ \hline
    Autoencoder (AE) &  Prefixes padded (PRFX) & One-hot encoding (OH) & Linear Temporal Logic (LTL) & Time since previous event (TSP) & Activity Suffix (SFX) \\ \hline
    Deep feedforward network (DFNN) & N-gram (N-GRAM) & Pretrained embedding (P-EMB) & Attributes (ATTR) & Day of week (WK) & Next timestamp (NT) \\ \hline
    Differentiable Neural Computer (DNC) & Single event (SE) & Frequency based (FB) & Clustered attributes (C-ATTR) & Time since midnight (TMD) & Remaining time (RT) \\ \hline
    Gated Recurrent Unit (GRU) & Timed state sample (TSS) & & Process model (PM) & Decayed time since previous place activation (DTPA) & Outcome (OUT) \\ \hline
    Bidirectional LSTM (B-LSTM) & & & Role (R) & & Resource (RES) \\ \hline
    Transformer (TRANS) & & & & & Role (ROLE) \\ \hline
    &&&&& Activity prediction in checkpoints (ACT $[$ CP $]$) \\ \hline
    &&&&& Role suffix (RLSFX) \\ \hline
    &&&&& Attribute suffix (ATTRSFX) \\ \hline

  \end{tabular}
  \caption{Morphological box of every possibility for each characteristic.}~\label{tab:morphological}
\end{table*}

\subsection{Input data}

The selection of input data used in a neural network is one of the most crucial decision to make when designing a predictive monitoring approach based on deep learning. In general, the most important information attributes in an event log, for the majority of prediction problems, would be the case identifier, the activities and the timestamp. If the prediction problem is an attribute or outcome, the required information will be dependent on the attribute in case.

Since most process logs used in the literature are real-life logs, these are prone to have noisy data. We refer to noisy data as a set of problems that includes missing data, duplicated tasks, inconsistent tasks, or other incorrect behaviour~\cite{Cheng2015}. Given that most logs used in predictive monitoring are real use cases, noisy logs are a prominent problem in process mining, even though most predictive monitoring approaches disregard this problem, or only try to minimize this problem by filtering traces using some condition. Missing data, as the absence of data values for a given variable in an observation~\cite{Graham2009}, is the most paradigmatic case of noise in process mining. This problem can take multiple dimensions. On the worst case, a task could be missing from the event log, so no information would be recorded in that case. On the best case, an event-level attribute could be missing for a given event. For example, the resource that executes the activity of an event could not be present. In the latter case, imputation techniques could be used directly over the process log to help improve the prediction performance~\cite{Osman2018}.

The time information available in the log requires some preprocessing to be exploited since it can not be used directly as inputs of the neural network. The most usual preprocessing applied to timestamps involves computing the difference between the timestamp of the current event and the timestamp of the previous event~\cite{Francescomarino2017,Tax2017,Navarin2017,Khan2018,Camargo2019,Wahid2019,Mauro2019}. Another used feature involves computing the difference between the current event timstamp and the timestamp from the beginning of the case~\cite{Francescomarino2017,Tax2017,Navarin2017,Khan2018,Wahid2019}. Other potentially useful features are the day of the week~\cite{Francescomarino2017,Tax2017,Navarin2017,Khan2018} or the time until the midnight~\cite{Francescomarino2017,Tax2017,Navarin2017,Khan2018,Wahid2019}. Additionally, the difference between the current timestamp and the timestamp of the last activation on a Petri net can be applied to a decay function which makes the calculated time decrease as the place is not activated~\cite{Theis2019}.

As far as the attributes of the log are concerned, it is possible that not every attribute adds significant
information to the predictive problem~\cite{Mehdiyev2017, Navarin2017, Schoenig2018, Mehdiyev2018, Lin2019, Hinkka2019, Theis2019, Wahid2019, Wang2019, Folino2019}. For example, and returning to the example in \tablename~\ref{tab:log}, if the e-mail information is present as part of an additional trace-level attribute, it will not be useful to predict the next medical test.~\cite{Lin2019} trains a predictive
model where an alignment weight vector learns the importance of each attribute.~\cite{Hinkka2019} clusters the attributes together using the x-means algorithm. The belonging to a cluster is added as an additional feature to the feature vector of each event. Furthermore, resources in a process log are
often noisy since some resources could potentially appear only once in the whole event log, even though they adhere to an
organizational scheme. To solve this problem,~\cite{Camargo2019} groups
resources depending on their activity execution profiles. Note that, as shown in \tablename~\ref{tab:studies} this is the only approach that distinguishes the resources from other attributes of the event log.
As far as the approaches that use time features~\cite{Francescomarino2017, Tax2017, Khan2018, Camargo2019, Theis2019, Wahid2019, Pasquadibisceglie2019, Mauro2019}, they face the problem of a high variability in the time between the events so these time features may complicate the training phase.

\subsection{Predictions}
Regarding the prediction targets, there are multiple possibilities:
\begin{itemize}
  \item \textit{Activity}~\cite{Evermann2017, Mehdiyev2017, Tax2017, Al-Jebrni2018, Khan2018, Schoenig2018, Mehdiyev2018, Camargo2019, Lin2019, Hinkka2019, Theis2019, Pasquadibisceglie2019, Mauro2019, Philipp2020}: the next activity of a running case.
  \item \textit{Activity suffix}~\cite{Francescomarino2017, Tax2017, Khan2018, Camargo2019, Lin2019}: the sequence of activities given a running case.
  \item \textit{Next timestamp}~\cite{Tax2017, Khan2018, Camargo2019}: the difference between the next timestamp of an event and
    the timestamp of the current event.
  \item \textit{Remaining time}~\cite{Tax2017, Navarin2017, Khan2018, Camargo2019, Wahid2019}: the difference between the timestamp of the last case of a trace and the current timestamp of an event.
  \item \textit{Outcome}~\cite{Metzger2018, Wang2019, Folino2019, Hinkka2019a}: refers to the outcome of a given running case.
  \item \textit{Attributes}~\cite{Camargo2019, Lin2019}: other event-level attributes present in the log. We include here the roles from \cite{Camargo2019} since they are derived from the resources of the log, which are, in turn, attributes of the log.
  \item \textit{Attribute suffix}~\cite{Camargo2019, Lin2019}: the approaches that predict the activity suffix and use other attributes of the log as inputs must predict also the sequence of attributes since the inputs for these attributes in the predicted events would be missing otherwise.
\end{itemize}

The most common prediction problem is the prediction of the next activity given a partial event prefix. When this is the case, other prediction tasks are taken up as auxiliary tasks that may help improve the prediction performance~\cite{Crawshaw2020}. The only exception is the remaining time prediction problem, since it can be approached as a direct predition, as shown in definition~\ref{def:remaining-time-direct}.

As far as the outcome prediction problem is concerned, most studies assume that the log
is fully labeled and, thus, the problem is treated in a supervised manner.
The exception to this is~\cite{Folino2019}, where they do not make such assumptions and, instead, consider
the log as a partially labeled dataset. Thus, the problem is treated in a
semi-supervised manner which has two stages: \textit{(i)} an LSTM is
trained on the full event log to predict the next activity, and \textit{(ii)} another
LSTM is initialized with the parameters learned by the previous model and it
is trained over the log that is labeled to predict the outcome of a running case.

When the objective is to predict the activity suffix from a partial trace, the next event must be sampled from
the output probability distribution of the last neural network's prediction layer. A
simple choice would be selecting the event with the highest probability in the
vector~\cite{Tax2017}. However, this solution often makes the predictions to be
very repetitive on the same top probability activity when the process is complex
enough. Even more, it could happen that the end of the trace is never predicted with
this method, which forces to add a maximum trace length constraint to avoid the
prediction to be infinite. Another solution is to perform a \textit{beam
 search} that explores a limited space of solutions by retaining the $b$ traces
that have the highest composed probability of events. This is the approach used
by~\cite{Francescomarino2017} in which the generation procedure stops when a
complete candidate trace conformant with a series of mined LTL rules is found.
They also decrease the probability of subsequent predictions of the same event.
In~\cite{Camargo2019}, another approach is taken, in which the next event is
sampled randomly following the probability distribution of the neural network.
Note that there exist more sampling methods, such as \textit{top-k
  sampling}~\cite{Fan2018} or \textit{nucleus sampling}~\cite{Holtzman2019}.

When predicting an activity suffix, every feature used as input, such as attributes or time features, must be predicted in subsequent steps.
These predictions are often approached using the same neural network to predict multiple features simultaneously.
Predicting multiple outputs at once in a neural network is also called
\textit{multitask learning}~\cite{Caruana1997}. It has been shown that predicting multiple tasks at once helps to enhance the generalization of the neural network because it increases the sample size used to train a deep learning model due to the extra information available by the results of multiple prediction tasks~\cite{Caruana1997}, i.e, assuming a hard parameter sharing model, the shared layers of the neural network would be affected by the gradients of all prediction tasks. Given that every prediction task is, by its nature, noisy, training a model for the main prediction task ideally should be able to not overfit that noise. For example, assuming that we are interested in forecasting the next activity, we could use the next timestamp and next resource prediction problems as auxiliary tasks. Since each task has a different noise pattern, a model that jointly learns multiple tasks would ``average'' those noise patterns, achieving thus a better representation of the main prediction task~\cite{Ruder2017}. This effect is especially relevant in predictive monitoring, where the
event logs are often scarce of data. Each
prediction task has attached its own loss, and the set of losses must be combined
so that they can be minimized. In predictive monitoring, the most usual form to combine the losses is
depicted in equation~\ref{loss-mtl}, where $L_t$ is the total combined loss,
$|T|$ is the total number of tasks, and $L_i$ is the loss of an individual task.

\begin{equation}
  \label{loss-mtl}
  L_t = \sum^{|T|}_{i=1} L_i
\end{equation}

This form of combining the task losses poses the problem that different tasks
could have different magnitudes, and one task
could dominate others just for a bigger value, which explains why sometimes multitask learning is avoided in predictive monitoring when the prediction target is only the next activity and not the activity suffix. Some solutions to this problem pose by manually setting a different weight for each of the tasks or by automatically learning the weights for those tasks~\cite{Cipolla2018}.

\subsection{Neural network types}

\subsubsection{Feedforward networks}
Feedforward networks~\cite{Goodfellow2016} are the most basic models for deep
learning. In this type of neural network, the information flows through it without any
recurrence. This lack of recurrence makes them well suited for
tabular data~\cite{Wahid2019} but not in process mining, where
event logs have dependencies between the activities 
that are not exploited by this type of network. Thus,
feedforward networks are not often used in predictive monitoring. However,
this type of neural network is commonly used in combination with other methods, such as autoencoders~\cite{Mehdiyev2017, Mehdiyev2018} or after extracting features from the process model~\cite{Theis2019}.

Every method described in this section uses an algorithm called \textit{backpropagation} which computes the gradients of a loss function with respect to the weights of the neural network. This allows the information from the loss function to flow ``backwards'' through the network to compute the gradient~\cite{Goodfellow2016}.

\subsubsection{Autoencoders}
The predictive monitoring approaches presented in~\cite{Mehdiyev2017} and~\cite{Mehdiyev2018} use autoencoders as their type of neural network.
This kind of neural network 
learns how to reconstruct its own input. An autoencoder has two main parts: the
encoder, which learns to map its input $x$ into a hidden representation $h$, and the
decoder, which learns to map a given hidden representation $h$ back into the original
input $x'$. The loss function is configured to penalize $x'$ from being
different from $x$. Formally, we would train an encoder $e$ and a decoder $d$ such as:

\begin{equation}
d(e(x)) = x'
\end{equation}

The autoencoders usually trained are \textit{undercomplete}, which means that the dimension of the hidden representation is less than the input dimension.
This forces the autoencoder to discern the most useful features of the input in its hidden representation.

\begin{figure}
  \centering
\includegraphics[width=0.35\textwidth]{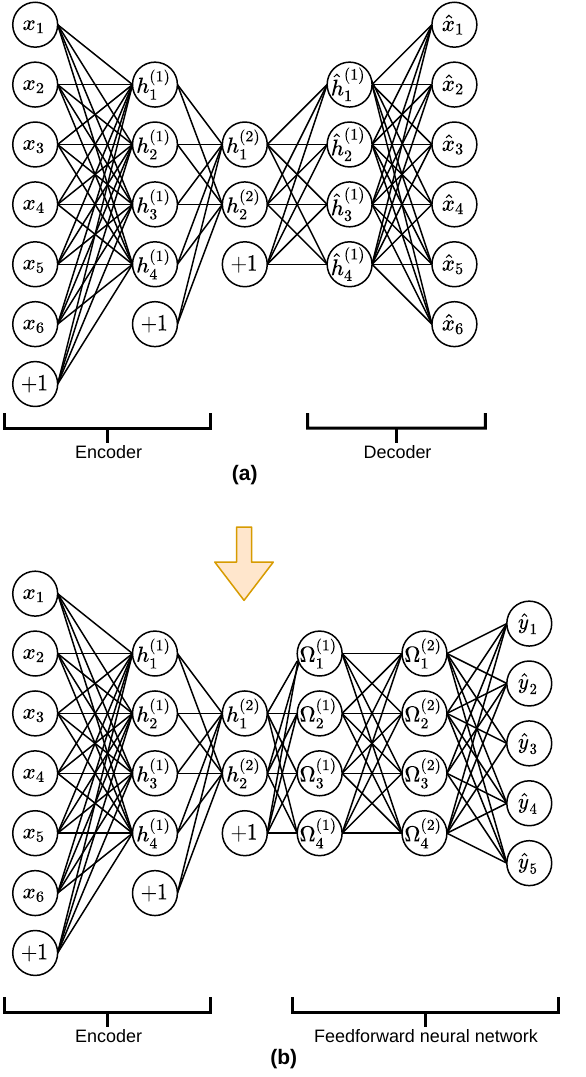}
\caption{Representation of the training of an autoencoder.}
\label{fig:autoencoder}
\end{figure}

Autoencoders may benefit from stacking multiple layers in the encoder and the
decoder in terms of representational power and computational complexity
reduction. Each layer reduces further the dimensionality of its input. The most
common way of training this type of autoencoder is by greedily feeding the
learned hidden features as the input of subsequent autoencoders. As shown in \tablename~\ref{tab:studies}, this approach is followed by~\cite{Mehdiyev2017, Mehdiyev2018} in which their inputs to the first
autoencoder are the hashed n-grams for each event prefix used to train the network. These approaches use the two-step procedure shown in \figurename~\ref{fig:autoencoder} to train an autoencoder. 
First, in \figurename~\ref{fig:autoencoder}a a stacked undercomplete autoencoder is trained to represent the most useful features by reducing the
dimensionality of its input. This autoencoder is composed of an encoder, which
maps the original input to a smaller hidden representation, and a decoder,
which maps the hidden representation back to the original input.
Then, in \figurename\ref{fig:autoencoder}b the already trained encoder is used to map the input to a
hidden representation. Then, a feedforward network ($\Omega$) is attached using the hidden
representation of the encoder as an input. In the final layer, a softmax
classification is applied to predict, in this case, the next activity.

In the context of predictive monitoring, an autoencoder approach may be useful when the number of attributes of the log is very high since it could help to reduce the dimensionality of the input data by selecting the most important features for each event. However, the main disadvantage of this approach resides in that they disregard longer depencencies between events of the partial trace.

The autoencoder architecture presented here could be improved by adding a
sparsity penalty to the loss function~\cite{Ranzato2006} or by corrupting the
inputs with noise~\cite{Alain2014}, but these improvements have not yet been explored
in predictive monitoring.

\subsubsection{Recurrent Neural Network}
\begin{figure}
  \centering
  \includegraphics[width=0.5\textwidth]{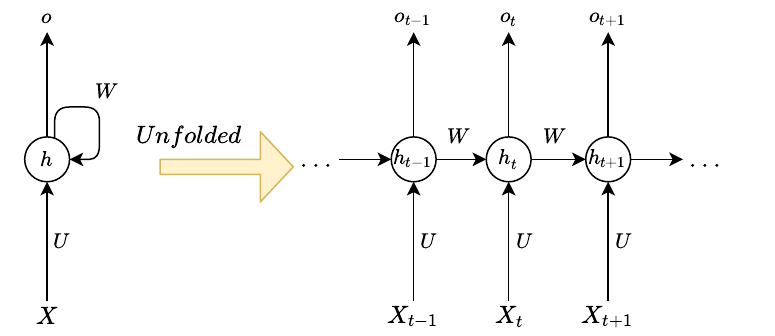}
  \caption{Graphical representation of a vanilla recurrent neural network. In
    the left part, the general model is presented. In the right part, the
    computational graph is unfolded three timesteps.}
  \label{fig:recurrence}
\end{figure}

Many deep predictive monitoring approaches~\cite{Evermann2017, Evermann2017a, Francescomarino2017, Tax2017, Navarin2017, Metzger2018, Schoenig2018, Camargo2019, Lin2019, Folino2019, Hinkka2019a, Hinkka2019, Hinkka2019a, Wang2019, Khan2018, Philipp2020} are based on Recurrent Neural Networks (RNN)~\cite{Goodfellow2016}.
RNNs are neural networks specialized in processing
sequential data, that is, they operate with a sequence of vectors $x_1,\ldots,
x_\tau$, where $\tau$ is the sequence length. The ability of processing sequence-like data makes this type of neural networks very useful to predictive monitoring.
\textbf{Simple Recurrent Neural Networks.}
In their simplest form, a recurrent neural network is graphically represented in \figurename~\ref{fig:recurrence}. It can be formally defined as follows:

\begin{equation}
  \label{rnn}
  \begin{aligned}
  & h_t = \tanh(b + Wh_{t-1} + Ux_t) \\
  & o_t = c + Vh_t
  \end{aligned}
\end{equation}

where:
\begin{itemize}
  \item $h_t$ denotes the hidden state of the recurrent neural network. This
    hidden state acts as an summary of the past sequence inputs up to the
    timestep $t$.
  \item $x_t$ refers to the input vector in the timestep $t$.
  \item $b$ and $c$ are bias vectors and $W$, $U$ and $V$ are weight matrices.
    The parameters of these matrices are updated with an algorithm called
    \textit{backpropagation through time} (BPTT)~\cite{Mozer1989AFB}, which allows applying the backpropagation algorithm to RNNs.
  \item $o_t$ is the output of the recurrent neural network in the timestep $t$.
\end{itemize}

This implementation of the RNN poses an important problem: the
gradients propagated using BPTT either vanish or explode when trying to learn
long dependencies. There exists multiple alternative models that have
been proposed to alleviate this problem: Long Short-Term Memory (LSTM), Gated Recurrent Unit (GRU) and Memory Augmented Networks (MANN).

\textbf{LSTM and GRU.}
As shown in \tablename~\ref{tab:studies}, LSTMs~\cite{Evermann2017, Evermann2017a, Francescomarino2017, Tax2017, Navarin2017, Metzger2018, Schoenig2018, Camargo2019, Lin2019, Folino2019, Hinkka2019a} and GRUs~\cite{Hinkka2019, Hinkka2019a} have been widely applied
in predictive monitoring and are two of the most popular architectures in this field. 
LSTMs~\cite{Gers1999} and GRUs~\cite{Cho2014} create paths through time that
allow the gradients to flow deeper in the sequence than in a vanilla RNN. Thus, instead
of using the previous state directly, $h_{t-1}$, LSTM, and GRUs use a memory cell
$C_t$ that has an internal recurrence and the usual recurrence of
vanilla RNN.

In the case of LSTMs, this internal recurrence is controlled by three different
gates, $f_t$, $o_t$, and $i_t$, which control the flow of information inside the cell.
$f_t$ is called the ``forget gate'', which filters what information is thrown away
from the cell state; $i_t$ is the ``input gate'', which controls what
information is going to be updated; and $o_t$ is the ``output gate'', 
which decides what information is exposed from the cell.
The definition of the formulas that define an LSTM is as follows:

\begin{equation}
f_t = \sigma(b_f + U_fx_t + W_fh_{t-1})
\end{equation}
\begin{equation}
i_t = \sigma(b_i + U_ix_t + W_ih_{t-1})
\end{equation}
\begin{equation}
o_t = \sigma(b_o + U_ox_t + W_oh_{t-1})
\end{equation}
\begin{equation}
\tilde{C_t} = tanh(b_C + U_Cx_t + W_Ch_{t-1})
\end{equation}
\begin{equation}
C_t = f_t \circ C_{t-1} + i_t \circ \tilde{C_t}
\end{equation}
\begin{equation}
h_t = o_t \circ tanh(C_t)
\end{equation}

In the previous equations, $x_t$ represents the input to the LSTM in the
timestep $t$, $b$ is a bias vector; $U$ and $W$ are trainable weight matrices;
$h_{t-1}$ represents the previous hidden state; $\tilde{C_t}$ is the calculation
of the cell state for the current timestep; and, finally $C_t$ is the combination of the past
information of the cell with the current information of the cell. The $\circ$
operation denotes the Hadamard product between two matrices.

GRUs are similar to LSTMs with the main difference that they do not have an
output gate. Formally, GRUs can be formally defined as follows:

\begin{equation}
  z_t = \sigma(W_zx_t + U_zh_{t-1} + b_z)
\end{equation}
\begin{equation}
  r_t = \sigma(W_rx_t + U_rh_{t-1} + b_r)
\end{equation}
\begin{equation}
  \tilde{h_t} = tanh(W_h x_t + U_h(r_t \circ h_{t-1}) + b_z )
\end{equation}
\begin{equation}
  h_t = z_t \circ h_{t-1} + (1 - z_t) \circ \tilde{h_t}
\end{equation}

In the previous equations, $z$ refers to the ``update gate'', which controls the
amount of information that flows from the past to the future; $r$ is called the
``reset gate'', which filters how much information from the past is forgotten;
$\tilde{h}$ represents the calculation of the current memory; and $h_t$
corresponds to the final calculation of the memory of the cell, which can be
interpreted as how much information is retained from the past and how much
information is updated.


In practice, there is almost no difference in the
performance of GRUs against LSTMs, but the former has the advantage of a faster
training~\cite{Chung2014, Wenpeng2017}.

\textbf{Bidirectional Recurrent Neural Network.}
\cite{Wang2019} uses a bi-directional LSTM with an attention mechanism to predict the outcome of a running case. Bi-directional recurrent neural networks, such as~\cite{Wang2019}, consist on applying one RNN forward and another RNN backwards, concatenating their hidden states of each timestep. Moreover, the attention mechanism, originally devised in~\cite{Luong2015} and \cite{Bahdanau2014}, allows learning an alignment vector to weight the importante of each timestep in the prediction.

\textbf{Memory Augmented Neural Network.}
In the predictive monitoring approach of~\cite{Khan2018}, an architecture that belongs to the family of the Memory Augmented Neural Networks (MANN) is proposed. 
The family of MANN architectures may be useful in predictive monitoring for learning longer dependencies when the traces of the log are very long or when cycles of the same event may make the LSTM and GRU to ``forget'' events in the beginning of the trace. However, these architectures are expensive to train and often very sensitive to the hyperparameters used.
MANNs use an external memory unit to enhance
the learning of longer term dependencies in sequences. The controller is often a Feedforward or a RNN that reads the inputs and, with the help of data from the memory, produces the corresponding outputs.

The oldest MANN architecture is the Neural Turing Machines~\cite{Graves2014} (NTM).
Instead of depending on a single cell
for having information from the past, the NTMs use an
addressing mechanism to access to this external memory cells. This addressing is
based on an attention mechanism that provides a weight vector, $w$,
which highlights the region of the memory more relevant for reading or
writing at each timestep. This addressing mechanism allows the neural network to both
interact with contiguous regions of memory and jump random addresses. One
possible implementation of the memory addressing would use as keys the internal state of the controller
LSTM in a certain timestep~\cite{Santoro2016}.
This kind of neural network is able to learn longer-term dependencies than its LSTM counterpart.
However, the NTMs suffer from training issues (slow convergence, NaNs in gradients, etc.). The Differentiable Neural Computer~\cite{Graves2016} (DNC) further improves the memory management of the NTM by allowing freeing allocated blocks, keeping track of the writes in memory and avoiding overlapping between memory blocks.

\begin{figure}
  \centering
  \includegraphics[width=0.5\textwidth]{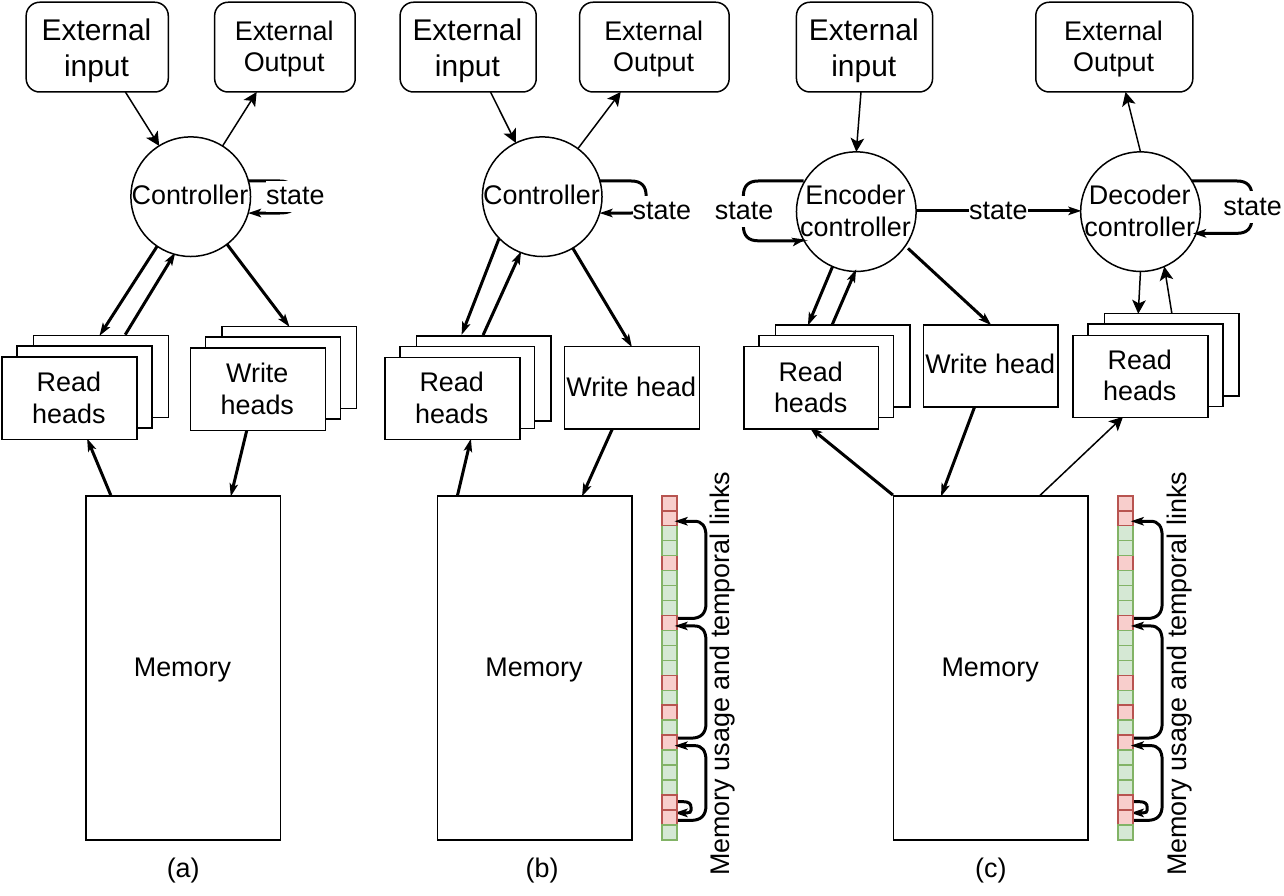}
  \caption{Comparison between the Neural Turing Machine~\cite{Graves2014} (\figurename~\ref{fig:dnc}a) the
    Differentiable Neural Computer~\cite{Graves2016} (\figurename~\ref{fig:dnc}b) and the variation of the
    DNC proposed by Khan et al. \cite{Khan2018} (\figurename~\ref{fig:dnc}).}~\label{fig:dnc}
\end{figure}

In~\cite{Khan2018}, a variation of the DNC is proposed. In this architecture, the
controller is separated into two controllers, the encoder controller and
the decoder controller, where both controllers are LSTMs.
The encoder reads the input event prefix
reading and writing the contents of the memory when necessary.
Then, the decoder is initialized with the last state of the LSTM
encoder, and the suffix is then predicted.
One notable aspect of this architecture is that the decoder is prevented from
writing into memory, so this architecture is write-protected.
\figurename~\ref{fig:dnc} shows a comparison between the NTM
from~\cite{Graves2014}, the DNC implementation from~\cite{Graves2016}, and the DNC implementation from~\cite{Khan2018}.

\textbf{Transformers.}
The predictive monitoring approach of~\cite{Philipp2020} uses a different architecture named the \textit{transformer}.
These type of architecture, originally proposed in~\cite{Vaswani2017}, substitutes the recurrence entirely by attention mechanisms. They rely on performing an attention operation over different parts of the sequence simultaneously (multi-head attention). In~\cite{Philipp2020} , instead of training a encoder-decoder like in the original proposal of the architecture, they only use the decoder part of the Transformer. Even though the transformer allows a faster training and inference due to the usage of only attention modules, it is still unclear how would the transformer deal with multiple heterogeneous input data, that is, when the inputs to the transformer model are both categorical and continuous data, such as the resources or time-related measures from the events.

\subsubsection{Convolutional neural network}
Convolutional Neural Networks were applied in the approaches of~\cite{Mauro2019, Al-Jebrni2018} and \cite{Pasquadibisceglie2019}, as shown in \tablename~\ref{tab:studies}.
This type of neural network is specialized in processing grid-like data. Most CNN applied to sequence prediction problems process the data as if it were an one-dimensional (1D) grid~\cite{Mauro2019, Al-Jebrni2018}.
In contrast, some CNNs reengineer their
preprocessing of the sequences to adapt them to a two-dimensional (2D)
grid~\cite{Pasquadibisceglie2019}. The two main operations performed by this type of
neural network are the \textit{convolution} and the \textit{pooling} operations.

The convolution operation takes two different arguments: \textit{(i)} the input
to the convolution operation, and \textit{(ii)} the \textit{kernel} (also called filter), which is a
matrix of learnable parameters much smaller than the input data. The output
of the convolution operation is called the \textit{feature map}. The convolution
operation slides the kernel through the input grid accross the input grid's width and height.
More formally, the usual convolution operation for 2D data is defined as follows~\cite{Goodfellow2016}:

\begin{equation}
  S(i, j) = (I * K)(i, j) = \sum^M_m \sum^N_n I(i + m, j + n)K(m, n)
\end{equation}

\begin{figure}
  \centering
  \includegraphics[width=0.5\textwidth]{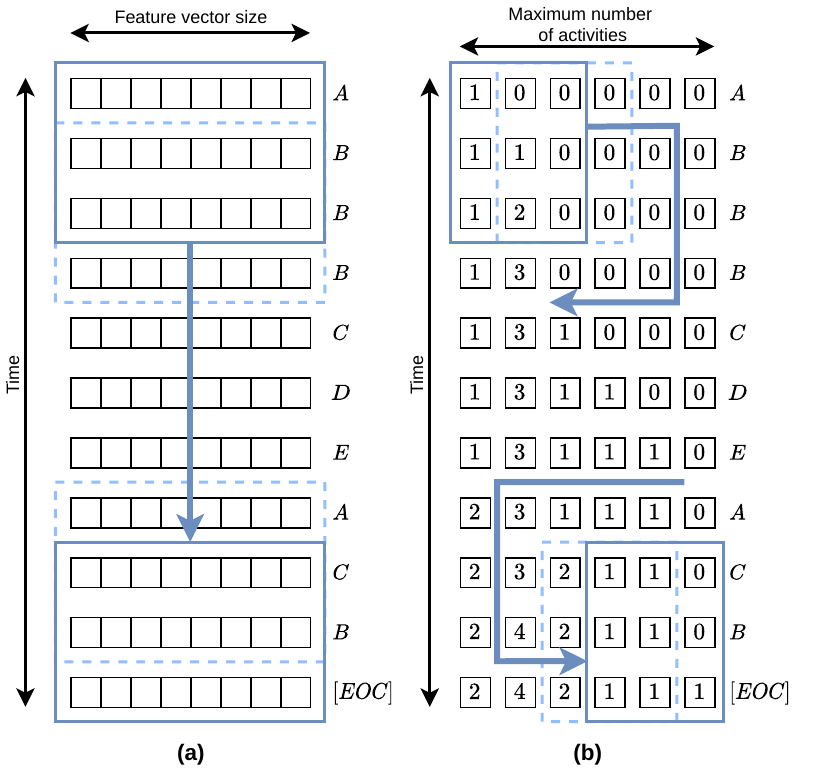}
  \caption{Differences between 1D convolutions~\cite{Al-Jebrni2018, Mauro2019}
    (\figurename~\ref{fig:convolution}a)
    and 2D convolutions~\cite{Pasquadibisceglie2019}
    (\figurename~\ref{fig:convolution}b) for processing the trace
    $ABBBCDEACB[EOC]$ assuming a total of 6 different activities, including
    the end of case, on the whole log. The blue square represents the filter and
    the blue arrow represents the direction of movement over the input matrix. For 1D convolutions, the filter is slided
    over the time dimension using the full width of the feature vector
    corresponding to the activities. For 2D convolutions, the filter is slided
    from right to left and from top to bottom using a smaller kernel size.}~\label{fig:convolution}
\end{figure}

Where $S(i,j)$ is the element $(i, j)$ of the feature map, $*$ is the
convolution operation, $M$ is the total height of the kernel, $N$ is the total
width of the kernel, $I$ is the input to the convolution operation, and $K(m, n)$
refers to the element $(m,n)$ of the kernel. In the case of 1D data, the convolution operation is applied over the tensor full width, which is the feature vector size, and with a certain height
(which is the time axis of the tensor, as shown in \figurename~\ref{fig:convolution}a.

\figurename~\ref{fig:convolution} highlights the differences between a 1D
convolution and a 2D convolution. Note that for a 2D convolution the time
dimension has to be added as an additional channel of the input, while for the
1D convolution, it would just enlarge the feature vector of each event. Thus,
for 2D convolutions, including information about the attributes of the event log requires
adding additional channels, which might pose a problem when the number of
different attributes is mismatched from the number of different activities (the
width of the matrix would not be the same).

After a convolution layer, the most usual next layer is the pooling layer, which applies a statistical summary of its output,
often the maximum or average of certain portions of the input grid. 
This kind of transformation has the advantage of learning invariant features to
the position inside the trace (local translation invariance). Furthermore,
applying a pooling operation can further reduce the dimensionality of the
problem, thus speeding up the training procedure.

The most usual basic architecture for a CNN is a series
of building blocks. Each block has a convolutional layer followed by a
non-linear activation function, and a pooling layer. The
works~\cite{Al-Jebrni2018, Pasquadibisceglie2019} are based on this neural architecture.~\cite{Al-Jebrni2018} follows the 1D convolutional approach, using embeddings for the
log categorical variables, whereas~\cite{Pasquadibisceglie2019} uses the
2D convolutional approach, using a frequency-based encoding for continuous and
discrete variables, as shown in \figurename~\ref{fig:convolution}b. In both studies, the non-linear activation function used
between layers is the ReLU function, which allows a faster training and alleviates the vanishing gradient problem~\cite{Krizhevsky2017}.

There are two ways to increase the expressivity of a neural network: increasing
its depth or its width. When increasing the depth the vanishing gradient problem arises: the gradient updates in
latter layers of the network are too small and impede the network to learn properly~\cite{He2016}.
When increasing the width, the number of
parameters, and therefore, its computational complexity, grows very rapidly.
This latter problem has been tackled
in~\cite{Szegedy2015}, where multiple modules of the neural perform both a
convolutional operation with different kernel sizes and a pooling operation simultaneously. This approach is used by~\cite{Mauro2019} but with two main differences: they use 1D
convolutions, and they do not perform a $1 \times 1$ convolution before applying a
convolution with a bigger size.

Comparing CNNs against RNNs for predictive monitoring, the former may have the advantage of a faster training and inference, specially for events logs with longer traces. However, RNNs may capture longer depencencies between the events of the trace since the hidden state for a given event depends on every event before it, whereas on a CNN it only depends on the $k$ most recent events, where $k$ is the size of the kernel.

\subsection{Sequence encoding}
In deep learning architectures, traces must be encoded in tensors of fixed size. However, there is a big variability in the length of every trace in a business process, so this step poses an important challenge. Furthermore, this step also conditions how the training targets are fed to the neural network. We have identified the following encoding formats:

\begin{itemize}
  \item \textit{Continuous}~\cite{Evermann2017, Al-Jebrni2018, Lin2019, Hinkka2019a, Philipp2020}: this encoding technique is inspired by the training of neural language models~\cite{Mikolov2010}. Here, the log is viewed as a text, each trace as a sentence of that text, and each activity as a word of a sentence. In this type of encoding, only a window $W$ of events is considered, and each window can include events from different traces. In case a window is incomplete, it could either be discarded or padded with zeroes. For example, let $L$ be the set of traces $\{[A,B,C], [C,D]\}$.  Then, with a window size $W = 3$ we would have the set of windows $\{[A,B,C], [EOC,C,D], [EOC,0,0]\}$ as an input. In this case, the training targets fed to the neural network are, in each timestep, the same set of windows shifted one position to the left, i.e, $\{[B,C,EOC], [C, D, EOC], [0, 0, 0]\}$.
  \item \textit{Prefixes padded}~\cite{Francescomarino2017, Tax2017, Navarin2017, Khan2018, Metzger2018, Camargo2019, Hinkka2019, Pasquadibisceglie2019, Wang2019, Mauro2019, Folino2019}: in this type of encoding, every possible set of event prefixes $hd^k$ where $k \in (1,...,n)$ for each trace is considered. There are two different approaches to apply this encoding. The first one considers only the $W$ most recent events (as in \cite{Camargo2019}). The second one considers all the events (as in \cite{Tax2017}). In both approaches, the event prefixes must be padded with zeroes in case they are shorter than the specified vector length. In the second case, the vector length is often set to the length of the longest trace of the log. The approaches that use this encoding set the training target for each event prefix to the next event that follows in the event prefix, even though the full event suffix of events could also be the training target.
  \item \textit{N-gram}~\cite{Mehdiyev2017, Mehdiyev2018}: this encoding, used by~\cite{Mehdiyev2017}, represents each trace as a set of all subsequences up to length $k$ contained in it. The total number of possible sequences for an event log of $|A|$ distinct activities can be calculated using equation \ref{ngram-calculation}.
    \begin{equation}
      \label{ngram-calculation}
      N = \sum^k_{i=1} |A|^i
    \end{equation}

    Since the space of n-gram combinations is very large, the ``hashing trick''
    \cite{Weinberger2009} (also known as ``feature hashing'') is used to reduce
    this dimensionality to a fixed length vector. The hashing trick is defined
    as in equation~\ref{hashing-trick}:
    \begin{equation}
      \label{hashing-trick}
      n_i = \sum_{i: h(i)=k} \xi(i)x_i
    \end{equation}

    A hash function $h$ is used to
    determine the $k$ position in the fixed vector that has to be updated with a
    feature $x_i$. Another hash function $\xi$ counters the effect of hash collisions
    by determining the sign of the update.

    In predictive monitoring, this type of encoding has only been used for autoencoders~\cite{Mehdiyev2017, Mehdiyev2018} so, in these proposals the training targets as well as the inputs of the NN are equal.

  \item \textit{Single event}~\cite{Wahid2019}: in this encoding, only a single event and its attributes
    are considered, so the sequence of events in the trace is disregarded. The approach of~\cite{Wahid2019} uses this encoding, setting the training targets as the next event to the event in question.

  \item \textit{Timed state}~\cite{Theis2019}: recently proposed by~\cite{Theis2019}, this encoding represents the inner state of a Petri net after replaying a partial trace in it. Each place of the Petri net is enhanced with a ``decay function'' that counts the time between the current timestamp and the last time a token was in a given place. The sequence encoded vector is defined as a concatenation of the following hand-crafted vectors: $F_t$ gives the value of the decay function for each place of the Petri net; $C_t$ counts the number of times a token has gone through a place of the Petri net; $M_t$ counts where the tokens are in the petri net, and $R_t$: counts the occurrence of other attributes of the trace. The training target is the next event after the replayed activity prefix on the Petri net.

The most used encodings are \textit{Continuous} and \textit{Prefixes padded}, since they are versatile enough to be used with both CNNs and RNNs. The \textit{single event} encoding is used less since it disregards the dependencies between the events of the log. The \textit{timed state} encoding has the advantage of using the model as an input and, potentially, to capture dependencies between the activities that are not present by examining the literal ordering of the events in the log. However, this encoding is not directly compatible with more expressive models such as RNNs or CNNs. In the comparison between \textit{continuous} and \textit{prefix padded} encodings, the former may benefit of a faster training at the expense of not capturing longer dependencies than in the latter encoding.
\end{itemize}

\subsection{Event encoding} 
Attribute variables can be either categorical variables or continuous variables. On the one hand, continuous variables must be normalized before feeding to the neural network. There are multiple techniques
such as log-normalization~\cite{Camargo2019}, min-max normalization~\cite{Camargo2019}, z-score normalization~\cite{2009a} or tanh-estimators~\cite{Scheirer2010}.
On the other hand, each categorical variable must be encoded in fixed feature
vectors that uniquely represents them. There are various strategies used in the
literature for that:
\begin{itemize}
  \item \textit{One-hot}~\cite{Francescomarino2017, Tax2017, Navarin2017, Khan2018, Metzger2018, Schoenig2018, Hinkka2019, Wang2019, Folino2019, Hinkka2019}: categorizing the variable with an integer is not
    enough since this categorization assumes that the higher the value of the
    variable is, the more important it is. To avoid that problem, the feature is
    represented in a binary vector where its size corresponds to the number of
    possible distinct values for that variable, and its position in the vector is
    a one if the category corresponds with the variable.

  \item \textit{Embedding}~\cite{Evermann2017, Al-Jebrni2018, Lin2019, Wahid2019, Mauro2019, Folino2019, Philipp2020}: the embedding
    encoding creates a matrix $W \in \mathbb{R}^{n \times f}$ where each row corresponds to each of the
    categories of the variable, and columns correspond to the feature dimension.
    The parameters of this matrix can be either established randomly or be
    learned with stochastic gradient descent, so the
    learned embeddings are optimal for the prediction task at hand.

  \item \textit{Frequency-based}~\cite{Theis2019, Pasquadibisceglie2019}: this type of encoding~\cite{Leontjeva2015} indicates how many times the activity $i$ has
    happened until the current event of the trace. This encoding is useful when
    temporal information must be added to the encoding of the activities. Note
    that in the case of \cite{Theis2019} the frequency does not represent
    directly activities but the number of times a token has gone through a place.

  \item \textit{Pretrained embedding}~\cite{Camargo2019}: instead of directly training the embedding vectors
    with stochastic gradient descent, the embeddings can be pretrained for
    another task that gives additional information. For example, in~\cite{Camargo2019}, a neural network is trained to learn embeddings by discriminating role-activity pairs. A positive instance means that a role-activity pair exists on the log whereas a negative pair means otherwise. They claim that the embeddings learnt this way can discriminate better between different events.
  \end{itemize}
  
   While the \textit{pretrained embedding} and \textit{embedding} approaches have the main purpose of learning a set of embeddings to represent the categorical variables, the former are richer and may provide more information, which eases the convergence of the neural network, as many works in NLP that used similar techniques have shown~\cite{Mikolov2013, Peters2018}. Moreover, the \textit{one-hot} encoding is a good solution in terms of easiness of computation and implementation, because it does not use additional parameters, otherwise, the size of the vector could dramatically increase the memory usage of the neural network. Note that associating an integer to a category and using that directly as an input is heavily discouraged, due to imposing an implicit order of the categorical variable that it is not true (i.e, category number 2 would be more important than category number 1)~\cite{Hancock2020}.

\section{Benchmark}
\subsection{Experimental setup}
\subsubsection{Datasets}
We performed the experiments using 12 real-life event logs from a variety of domains\footnote{The adaptations made to the approaches, the original code repositories and the attributes used by the approaches are available in the supplementary material.}. These event logs
were extracted from the \textit{4TU Center for Research
  Data}\footnote{The logs are available in the following repository: \url{https://github.com/ERamaM/PredictiveMonitoringDatasets}}
and are also available in the repository of our comparison tool.
\tablename~\ref{tab:log-info} shows some relevant statistics from these logs, namely, the number of cases, the number of different activities, the number of events, the average and maximum case length, the maximum and average event duration in days, the average and maximum case duration in days and the number of different variants. Most logs have a high event time variability (difference betweeen average event duration and maximum event duration), and a high trace length variability. The log ``Nasa'' shows 0 in the time related measures since the time variability in this log is low.

\begin{table*}[htbp]
  \scriptsize
  \centering
\begin{tabularx}{0.8\linewidth}{|p{2.8cm}|L|L|L|L|L|L|L|L|L|L|}
  \hline
Event log & Num. cases           & Num. activities      & Num. events & Avg. case length & Max. case length & Avg. event duration & Max. event duration & Avg. case duration & Max. case duration & Variants \\ \hline
Helpdesk & 4580 & 14 & 21348 & 4.66 & 15 & 11.16 & 59.92 & 40.86 & 59.99 & 226
\\
BPI 2012 & 13087 & 36 & 262200 & 20.04 & 175 & 0.45 & 102.85 & 8.62 & 137.22 &
4366 \\
BPI 2012 Complete & 13087 & 23 & 164506 & 12.57 & 96 & 0.74 & 30.92 & 8.61 &
91.46 & 4336 \\
BPI 2012 W & 9658 & 19 & 170107 & 17.61 & 156 & 0.7 & 102.85 & 11.69 & 137.22 &
2621 \\
BPI 2012 W Complete & 9658 & 6 & 72413 & 7.5 & 74 & 1.75 & 30.92 & 11.4 & 91.04
& 2263 \\
BPI 2012 O & 5015 & 7 & 31244 & 6.23 & 30 & 3.28 & 69.93 & 17.18 & 89.55 & 168
\\
BPI 2012 A & 13087 & 10 & 60849 & 4.65 & 8 & 2.21 & 89.55 & 8.08 & 91.46 & 17
\\
BPI 2013 closed problems & 1487 & 7 & 6660 & 4.48 & 35 & 51.42 & 2254.84 &
178.88 & 2254.85 & 327 \\
BPI 2013 incidents & 7554 & 13 & 65533 & 8.68 & 123 & 1.57 & 722.25 & 12.08 &
771.35 & 2278 \\
Sepsis & 1049 & 16 & 15214 & 14.48 & 185 & 2.11 & 417.26 & 28.48 & 422.32 & 845 \\
Env. permit & 1434 & 27 & 8577 & 5.98 & 25 & 1.09 & 268.97 & 5.41 &
275.84 & 116 \\
Nasa & 2566 & 94 & 73638 & 28.7 & 50 & 0.0 & 0.0 & 0.0 & 0.0 & 2513 \\ \hline

\end{tabularx}
\caption{Statistics of the event logs used for benchmarking. Time related
  measures are shown in days.}~\label{tab:log-info}
\end{table*}

\subsubsection{Data split}


  We performed a 5-fold cross-validation in which each of the folds is used to report the final performance of the deep learning approach. For each remaining set of 4-folds, we split it in training and validation sets in a 80/20 ratio. These latter partitions are used to train the neural network and to select the best performing model, respectively. Each of the approaches is tested one time in each of the folds. We refrain from increasing the number of folds, training repetitions or performing a nested cross-validation for two reasons. First, some approaches~\cite{Khan2018} are computationally expensive to train or perform an hyperparameter optimization that heavily increases training time~\cite{Camargo2019,Mauro2019}. Second, and more importantly, the time to perform the testing phase for the suffix prediction tasks when the prediction is not direct (definitions 5, 7, and 9 from Section 2) is very high due to having to predict every single event for each possible prefix until the end of the case is reached or the maximum trace length is achieved. For some approaches~\cite{Tax2017} this testing time is orders of magnitude greater than the time required to train the neural network.

Furthermore, many approaches add an ``end of case'' token at the end of every trace of the log. This modification allows to give a clear stop condition for the next activity and activity suffix prediction problems. Thus, we do not have to rely on inspecting the set of last activities of every trace of the log to know whether a case has finished or not. Furthermore, in this latter case we would not get a definite answer either, because the last activity of the trace could be part of a loop, e.g., a case that always repeats the same activity. To unify the procedure and to make the metrics of next activity prediction comparable, we augmented the log with an end of case activity at the end of each trace for every approach tested.

\subsubsection{Metrics}
Depending on the predictive task, we use the following metrics for evaluating the performance of the approaches:

\begin{itemize}
  \item \textit{Next activity prediction}\footnote{More metrics for the next activity prediction task are available in the supplementary material.}: since the next activity prediction task is a classic classification problem, we use the \textit{accuracy} metric. The accuracy measures the proportion of correct classifications in relation to the number of predictions done. Other measures such as the Matthews Correlation Coefficient~\cite{Matthews1975} or the weighted F1 score are reported by the approaches tested but we found that the results were aligned with the accuracy measure and do not give additional information. Therefore, these results are not reported in this paper.

  \item \textit{Activity suffix prediction}: when predicting an activity suffix in the context of predictive monitoring, it is important to take into account that the activities in the process may occur in parallel~\cite{Tax2017}. Thus, instead of the metrics used for the next activity prediction task, we use the \textit{Damerau-Levenshtein distance} metric. This metric measures the edit distance between two given strings without penalizing too harshly transpositions of tokens, which, in the context of predictive monitoring, could mean a pair of parallel activities. These two strings represent the predicted activity suffix for a given event prefix and its ground truth activity suffix. The Damerau-Levenshtein metric measures the number of insertions, deletions, substitutions, and transpositions needed to transform one string into another. This value is then normalized by the lengths of the two strings, obtaining a value of similarity between $0$ and $1$.

  \item \textit{Next timestamp prediction and remaining time prediction}\footnote{The results of the next timestamp prediction problem are available in the supplementary material.}: since the time prediction problem is a regression task, the metric selected for measuring the performance in the next timestamp prediction tasks and remaining time prediction tasks is the \textit{Mean Absolute Error} (MAE). This metric is defined in equation~\ref{eq:mae} and  has the advantage of not overpenalizing the variability in the observations~\cite{Willmott2005}, which is the case in time prediction in predictive process monitoring, where the time between two events in a trace can be potentially large~\cite{Tax2017, Verenich2019}.
     \begin{equation}
       \label{eq:mae}
       MAE = \frac{\sum^N_{i=1} |y_i - \hat{y}_i|}{N}
     \end{equation}
  \end{itemize}

  Independently of the prediction problem and metric used, we apply a two stage statistical test to quantify the differences between the predictive monitoring approaches. Instead of applying some of the classical null hypothesis significance testing (NHST) approaches, we rely on Bayesian analysis to assess the differences between the approaches, mainly due to the following reasons: \textit{(i)} NHST yields no information about the null hypothesis, so, if the null hypothesisis is not rejected, we would not gather any valuable information of the statistical test~\cite{Benavoli2017}, \textit{(ii)} NHST do not estimate probabilities of the hypotheses, so comparing the classifiers would be more difficult~\cite{Benavoli2017}, and \textit{(iii)} as far as we know, there is no NHST that can take into account the variance of the results over multiple folds of a cross-validation testing procedure since those tests work by aggregating the results of the folds for each dataset~\cite{Demsar2006}. We rely on the library \texttt{scmamp}~\cite{Calvo2016} for performing these statistical tests.

  We propose a two stage statistical comparison procedure. First, we apply a Bayesian analysis technique based on Plackett-Luce model~\cite{Calvo2018} that enables ranking multiple algorithms at the same time, allowing to get an overall comparison and probabilities of being the best classifier for the whole set of tested predictive monitoring approaches. Since the Plackett-Luce approach does not take into account the individual fold results of the cross-validation testing procedure, and to try to solve ties between the approaches, we apply a hierarchical Bayesian model to pairwise compare the three best approaches in terms of Plackett-Luce ranking~\cite{Corani2017}.

\subsubsection{Approaches and experimental setup}

We performed the benchmark with 10 different approaches from the state of the art~\cite{Pasquadibisceglie2019,Mauro2019,Tax2017,Evermann2017,Khan2018,Theis2019,Navarin2017,Camargo2019,Hinkka2019,Francescomarino2017}. We discarded the remaining 11 surveyed approaches based on two conditions.

On the one hand, we decided to discard the four approaches that predict outcome because they are a special case of predictive monitoring where, unlike the other approaches analyzed in this paper, the solutions are focused on a particular set of logs. Thus, they often define different outcomes, adding additional information from the application domain. Since they are ad-hoc approaches for specific logs and outcomes, they are left out of the scope of this review.

On the other hand, we discarded the remaining seven approaches since they had no code publicly available. For these approaches, we contacted the authors but received no response or the code was no longer available or it could not be shared. Note that we also contacted the authors even when we had the code if we found blocking issues that prevented us to completely execute the code of the approach.

  As far as the hyperparameter optimization is concerned, we decided to apply the hyperparameter optimization procedure only if it is already available in their code. For the remaining approaches, we use the original hyperparameters, even though their performance could be further enhanced by an optimization procedure~\cite{Hutson2020}. In fact, only~\cite{Camargo2019} and~\cite{Mauro2019} optimize their hyperparameters using a random search and Hyperopt, respectively. The rest of the approaches were left with their default hyperparameters reported in the papers or in their source code.

The experiments were carried out in a server equipped with an Intel Xeon Gold 5220, 192 GB RAM, and 2 Nvidia Tesla V100S 32GB. Trace level attributes have not been taken into consideration. To perform the specific preprocessing for each dataset, we rely on the Pm4Py library~\cite{Berti2019}. As shown in the aforementioned table, the highest count of attributes is present in the datasets ``BPI 2013 Incidents'' and ``Nasa''. Note that not every dataset has available the resource assigned to the event so, if the approach uses specifically this attribute, it can not be evaluated (which is the case of Camargo et al.~\cite{Camargo2019} in datasests ``Nasa'' and ``Sepsis''). If an approach can not be evaluated in a dataset, this obligues to discard that dataset from the statistical comparison. Thus, the provided statistical comparisons are performed without taking into account the ``Nasa'' and ``Sepsis'' datasets.

\subsection{Results and discussion}

\tablename~\ref{tab:accuracy} shows the mean accuracy of the 5-fold cross-validation testing procedure for the next activity prediction problem. From a frequentist point of view, Hinkka et al.~\cite{Hinkka2019} obtains the best result in 5 process logs, Pasquadibisceglie et al.~\cite{Pasquadibisceglie2019} obtains the best result in one log, Tax et al.~\cite{Tax2017} obtains the best result in 4 logs and Theis et al.~\cite{Theis2019} obtains the best result in 2 logs.

\begin{table*}
\centering
\begin{tabular}{l|cccccccccccc}

{} &     \rotatebox{90}{BPI 2012} &   \rotatebox{90}{BPI 2012 A} & \rotatebox{90}{\shortstack[l]{BPI 2012 \\ Complete}} &   \rotatebox{90}{BPI 2012 O} &   \rotatebox{90}{BPI 2012 W} & \rotatebox{90}{\shortstack[l]{BPI 2012 \\ W Complete}} & \rotatebox{90}{\shortstack[l]{BPI 2013 \\ Closed Problems}} & \rotatebox{90}{\shortstack[l]{BPI 2013 \\ Incidents}} &             \rotatebox{90}{Env Permit} &               \rotatebox{90}{Helpdesk} &                   \rotatebox{90}{Nasa} &                 \rotatebox{90}{Sepsis} \\ \hline

Camargo           &                                  83.28 &                                  75.98 &                                       77.93 &                                  81.35 &                                   76.4 &                                         68.95 &                                              54.67 &                                        66.68 &        \textcolor{red}{\textbf{85.78}} &                                  82.93 &                                    - &                                    - \\ \hline
Evermann          &                                  59.33 &                                  75.82 &                                       62.38 &                                  79.42 &                                  75.37 &                                         67.53 &                                              58.83 &              \textcolor{red}{\textbf{66.78}} &                                  76.19 &        \textcolor{red}{\textbf{83.66}} &                                  20.37 &                                   40.0 \\ \hline
Hinkka            &  \textcolor{PineGreen}{\textbf{86.65}} &  \textcolor{PineGreen}{\textbf{81.19}} &       \textcolor{PineGreen}{\textbf{80.64}} &  \textcolor{PineGreen}{\textbf{87.23}} &                                  84.78 &               \textcolor{red}{\textbf{70.54}} &                 \textcolor{orange}{\textbf{63.47}} &        \textcolor{PineGreen}{\textbf{74.69}} &                                  84.43 &                                  83.08 &        \textcolor{red}{\textbf{88.42}} &      \textcolor{orange}{\textbf{63.5}} \\ \hline
Khan              &                                   42.9 &                                   74.9 &                                       47.37 &                                  66.08 &                                  60.15 &                                         52.22 &                                              43.58 &                                        51.91 &                                  83.59 &                                  79.97 &                                  12.71 &                                  21.01 \\ \hline
Mauro             &        \textcolor{red}{\textbf{84.66}} &     \textcolor{orange}{\textbf{79.76}} &             \textcolor{red}{\textbf{80.06}} &     \textcolor{orange}{\textbf{82.74}} &        \textcolor{red}{\textbf{85.98}} &                                         68.64 &                                              24.94 &                                        36.67 &                                  53.59 &                                  31.79 &                                  21.03 &        \textcolor{red}{\textbf{61.52}} \\ \hline
Pasquadibisceglie &                                  83.25 &                                  74.12 &                                        74.6 &                                  78.88 &                                  81.19 &                                         68.34 &                                              47.45 &                                        46.03 &  \textcolor{PineGreen}{\textbf{86.69}} &     \textcolor{orange}{\textbf{83.93}} &                                  88.27 &                                  56.15 \\ \hline
Tax               &     \textcolor{orange}{\textbf{85.46}} &        \textcolor{red}{\textbf{79.53}} &          \textcolor{orange}{\textbf{80.38}} &        \textcolor{red}{\textbf{82.29}} &                                  85.35 &                                         69.79 &              \textcolor{PineGreen}{\textbf{64.01}} &           \textcolor{orange}{\textbf{70.09}} &                                  85.71 &  \textcolor{PineGreen}{\textbf{84.19}} &  \textcolor{PineGreen}{\textbf{89.44}} &  \textcolor{PineGreen}{\textbf{64.22}} \\ \hline
Theis et al. (w/o attributes) &                                  82.89 &                                   65.5 &                                       75.26 &                                  78.38 &     \textcolor{orange}{\textbf{86.22}} &            \textcolor{orange}{\textbf{80.06}} &                    \textcolor{red}{\textbf{59.48}} &                                        59.41 &     \textcolor{orange}{\textbf{86.29}} &                                  78.77 &     \textcolor{orange}{\textbf{88.96}} &                                  55.74 \\ \hline
Theis et al. (w/ attributes)    &                                  80.96 &                                  65.67 &                                       75.75 &                                  76.89 &  \textcolor{PineGreen}{\textbf{86.86}} &         \textcolor{PineGreen}{\textbf{83.84}} &                                              54.65 &                                         51.5 &                                  85.12 &                                  79.69 &                                  86.34 &                                  58.14 \\ \hline

\end{tabular}
\caption{Mean accuracy of the 5-fold cross-validation for the next activity prediction problem. The best, second best and third best approaches are highlighted in green, orange and red, respectively.}~\label{tab:accuracy}
\end{table*}

\begin{figure}[ht]
  \centering
  \includegraphics[width=0.5\textwidth]{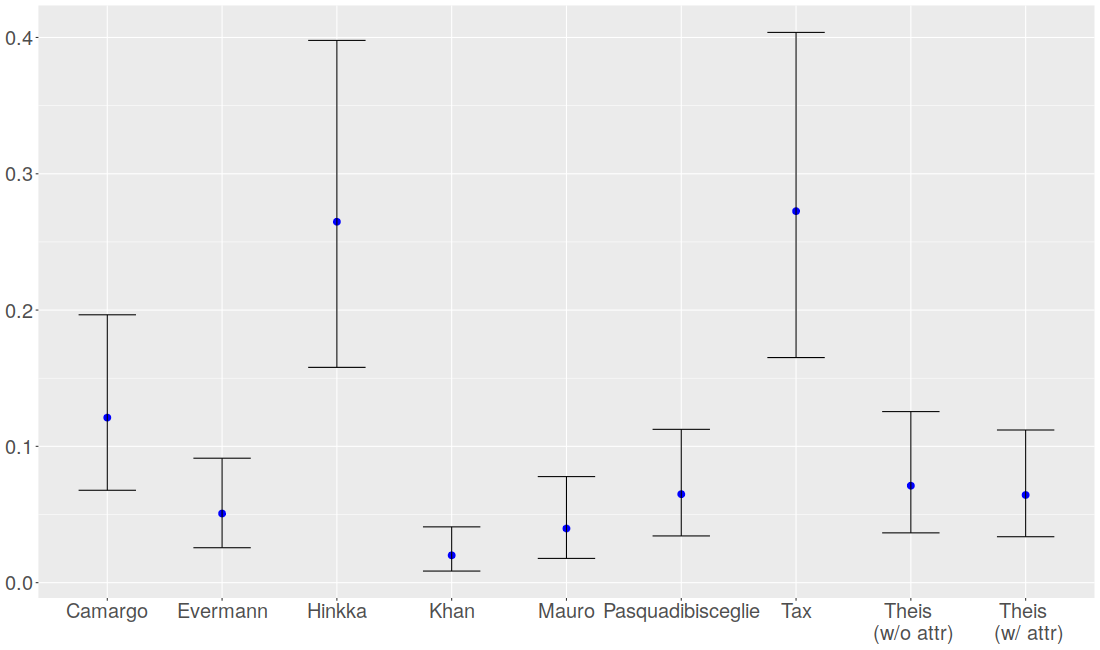}
  \caption{Credible intervals (5\% and 95\% quantiles) and expected probability of winning for the next activity prediction problem.}~\label{fig:ci-accuracy}
\end{figure}

From a Bayesian point of view, the best approach according to the Plackett-Luce rankings, shown in \tablename~\ref{tab:plackett-accuracy} is Tax et al~\cite{Tax2017}, followed closely by Hinkka et al.~\cite{Hinkka2019} and, at more distance, by Camargo et al.~\cite{Camargo2019}. \figurename~\ref{fig:ci-accuracy} shows the credible intervals calculated from the 5\% and 95\% quantiles alongside with the probability of winning. In this plot, if two approaches are not overlapped, they are statistically significant different. Thus, we can distinguish three blocks of approaches. The approaches of Hinkka~\cite{Hinkka2019} and Tax~\cite{Tax2017}, that are only overlapped by themselves and by Camargo et al. The approach of Camargo~\cite{Camargo2019}, which overlaps every other approach and the rest of approaches that only overlap themselves and Camargo.

\begin{table*}[ht]
  \centering
  \begin{tabular}{c|l|l|l|l|l|l|l|l|l|}
    Approach & Tax & Hinkka & Camargo & Theis (w/o resource) & Pasqua. & Theis (w/ resource) & Evermann & Mauro & Khan \\ \hline
    Ranking & 1.5288 & 1.5908 & 3.2906 & 5.1021 & 5.4615 & 5.5126 & 6.4451 & 7.2742 & 8.7934 \\ \hline
    Probability & 27.69\% & 26.96\% & 12.51\% & 7.47\% & 6.80\% & 6.74\% & 5.36\% & 4.27\% & 2.18\% \\ \hline
  \end{tabular}
  \caption{Plackett-Luce probability of winning and ranking for the next activity prediction problem.}~\label{tab:plackett-accuracy}
\end{table*}

\begin{figure*}[ht]
  \centering
  \subfloat[Camargo vs Hinkka]{\includegraphics[width=0.3\textwidth]{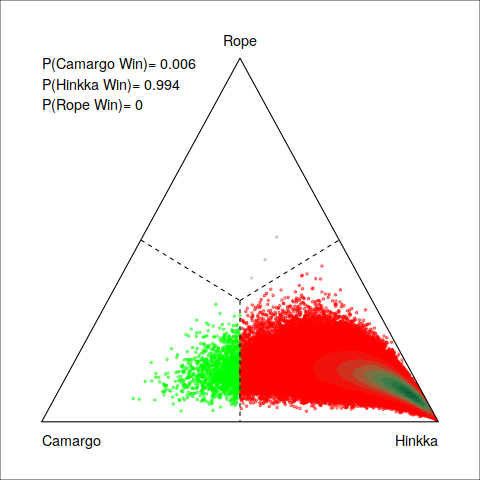}\label{fig:hie-a}}
  \qquad
  \subfloat[Camargo vs Tax]{\includegraphics[width=0.3\textwidth]{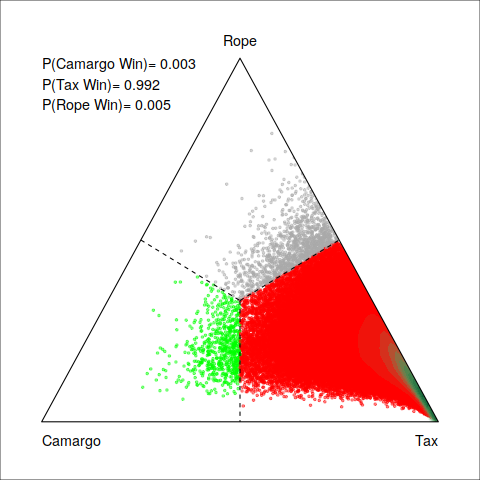}\label{fig:hie-b}}
  \qquad
  \subfloat[Hinkka vs Tax]{\includegraphics[width=0.3\textwidth]{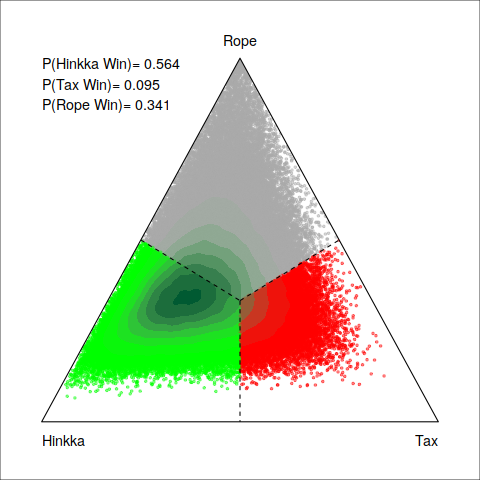}\label{fig:hie-c}}
  \caption{Hierarchical Bayesian tests for the next activity prediction problem.}~\label{fig:hierarchical-accuracy}
\end{figure*}

\figurename~\ref{fig:hierarchical-accuracy} shows the simplex plots of the hierarchical Bayesian tests. These plots show all the samples obtained from the posterior probability distribution of the approaches being equal, best or worse than the other. The closer the sample falls to each of the vertices, the more confidence we will have that this sample belongs to that region (and viceversa). Thus, the two bottom vertices of the triangle represent the pair of approaches being compared, the top vertex (rope) represents the case of the approaches being equal, and the dark green circle represents where the highest density of samples falls off. We have also included the probability of each region, denoted as $P(X \: win)$, where $X$ is the region under consideration.

The simplex plots of \figurename~\ref{fig:hierarchical-accuracy} show that the Hinkka~\cite{Hinkka2019} and Tax~\cite{Tax2017} outperform the approach of~\cite{Camargo2019} with a probability of 99.4\% and 99.2\% (\figurename~\ref{fig:hie-a} and~\ref{fig:hie-b}), respectively, and the density of the samples favours the former group of approaches. The approach of Hinkka~\cite{Hinkka2019} outperforms the approach of Tax~\cite{Tax2017} with a probability of 56.5\%, and they have a probability of being equal approaches of a 33.9\% (\figurename~\ref{fig:hie-c}). Thereby, the probability of the approach of Hinkka~\cite{Hinkka2019} of not being equal to the approach of Tax~\cite{Tax2017} is of 90.4\%. Surprisingly, none of the approaches that optimize hyperparameters (Mauro~\cite{Mauro2019} and Camargo~\cite{Camargo2019}) obtain any significant improvement. It is noteworthy to mention that there is an overall small difference between the Hinkka~\cite{Hinkka2019} approach against Tax~\cite{Tax2017} where the former approach promotes a heavy usage of the attributes of the log whereas the latter does not. Even though, it is true that Hinkka~\cite{Hinkka2019} outperforms Tax~\cite{Tax2017} in logs such as BPI 2012 O, the BPI 2013 Incidents or the BPI 2012 A, in other ones, such as the environmental permit, helpdesk, or Sepsis, they obtain worse results. Thus, this signals that, except in particular logs, it is to be expected that, overall, the usage of additional attributes may not be benefitial at all and even detrimental to the performance of the neural network.

As far as the suffix prediction problem is concerned, the best approach according to the Plackett-Luce rankings, shown in \tablename~\ref{tab:plackett-dl}, is the approach of Camargo~\cite{Camargo2019} with the random sampling procedure. The ``argmax'' procedure samples the next activity always as the one with the highest probability, while the ``random'' procedure samples the next activity following the probability distribution of the predicted activities. The ``argmax'' procedure is often stuck in repeating the same set of activities in logs with longer traces and loops due to selecting always the most probable set of activities whereas the ``random'' sampling has a better chance to escape these prediction repetitions~\cite{Holtzman2019}. Thus, the ``random'' sampling procedure is expected to outperform the ``argmax'' procedure in complex logs with long traces, where the chance of getting stuck predicting the same set of activities is low. However, in more simple logs, with shorter traces, the ``argmax'' procedure should slightly outperform the ``random'' procedure since the suffixes can be predicted with more precision.

\begin{figure*}[ht]
  \centering
  \subfloat[Camargo (argmax) vs Camargo (random)]{\includegraphics[width=0.3\textwidth]{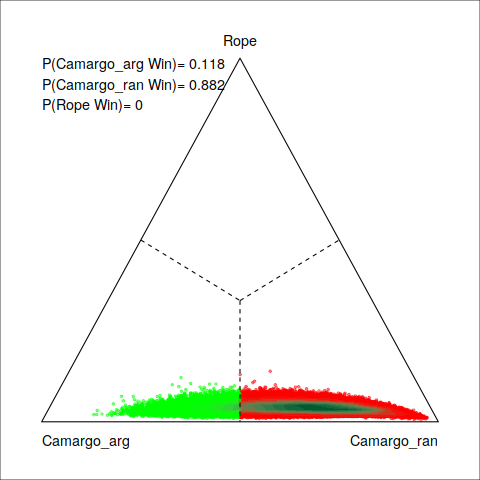}\label{fig:hie-a-dl}}
  \qquad
  \subfloat[Camargo (argmax) vs Evermann]{\includegraphics[width=0.3\textwidth]{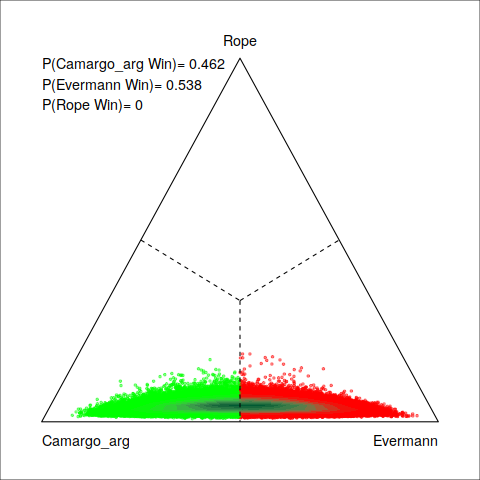}\label{fig:hie-b-dl}}
  \qquad
  \subfloat[Camargo (random) vs Evermann]{\includegraphics[width=0.3\textwidth]{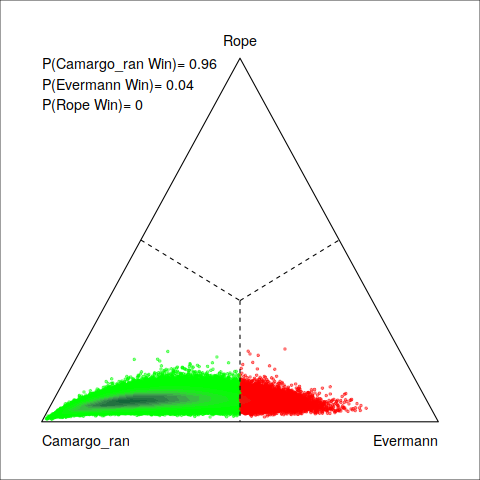}~\label{fig:hie-c-dl}}
  \caption{Hierarchical Bayesian tests for the suffix prediction problem.}~\label{fig:hierarchical-dl}
\end{figure*}

Nevertheless, the ``random'' procedure works reliably across a wide range of datasets, as shown in the hierarchical Bayesian tests of \figurename~\ref{fig:hie-a-dl} and~\ref{fig:hie-c-dl}. Note that the approach of Evermann~\cite{Evermann2017} works slightly better than the approach of Camargo~\cite{Camargo2019} and much better than the approach of Tax~\cite{Tax2017} suggesting that multitask learning does not benefit the activity suffix prediction tasks, specially when the auxiliary tasks are quite difficult, such as the time-related ones.

\begin{table*}[ht]
\centering
\begin{tabular}{l|cccccccccccc}

{} &      \rotatebox{90}{BPI 2012} &    \rotatebox{90}{BPI 2012 A} & \rotatebox{90}{\shortstack[l]{BPI 2012 \\ Complete}} &    \rotatebox{90}{BPI 2012 O} &   \rotatebox{90}{BPI 2012 W} & \rotatebox{90}{\shortstack[l]{BPI 2012 \\ W Complete}} & \rotatebox{90}{\shortstack[l]{BPI 2013 \\ Closed Problems}} & \rotatebox{90}{\shortstack[l]{BPI 2013 \\ Incidents}} &             \rotatebox{90}{Env Permit} &               \rotatebox{90}{Helpdesk} &                   \rotatebox{90}{Nasa} &                  \rotatebox{90}{Sepsis} \\ \hline

Camargo (argmax)  &        \textcolor{red}{\textbf{0.1851}} &  \textcolor{PineGreen}{\textbf{0.6536}} &            \textcolor{red}{\textbf{0.2218}} &  \textcolor{PineGreen}{\textbf{0.6845}} &       \textcolor{red}{\textbf{0.1941}} &                                        0.0501 &             \textcolor{PineGreen}{\textbf{0.6641}} &                                       0.2607 &  \textcolor{PineGreen}{\textbf{0.844}} &  \textcolor{PineGreen}{\textbf{0.911}} &                                    - &                                     - \\ \hline
Camargo (random)  &  \textcolor{PineGreen}{\textbf{0.3891}} &     \textcolor{orange}{\textbf{0.6441}} &      \textcolor{PineGreen}{\textbf{0.4495}} &     \textcolor{orange}{\textbf{0.6042}} &  \textcolor{PineGreen}{\textbf{0.311}} &           \textcolor{orange}{\textbf{0.3191}} &                                             0.5357 &       \textcolor{PineGreen}{\textbf{0.5294}} &       \textcolor{red}{\textbf{0.7595}} &       \textcolor{red}{\textbf{0.8524}} &                                    - &                                     - \\ \hline
Evermann        &     \textcolor{orange}{\textbf{0.1986}} &        \textcolor{red}{\textbf{0.5847}} &         \textcolor{orange}{\textbf{0.2693}} &        \textcolor{red}{\textbf{0.5544}} &      \textcolor{orange}{\textbf{0.28}} &        \textcolor{PineGreen}{\textbf{0.3372}} &                \textcolor{orange}{\textbf{0.6416}} &           \textcolor{orange}{\textbf{0.473}} &                                 0.5713 &                                 0.8354 &    \textcolor{orange}{\textbf{0.1218}} &  \textcolor{PineGreen}{\textbf{0.2693}} \\ \hline
Francescomarino &                                  0.1321 &                                  0.2871 &                                      0.0883 &                                  0.4591 &                                  0.141 &              \textcolor{red}{\textbf{0.1126}} &                                             0.5276 &             \textcolor{red}{\textbf{0.3607}} &                                 0.3924 &                                 0.4619 &       \textcolor{red}{\textbf{0.0842}} &        \textcolor{red}{\textbf{0.0742}} \\ \hline
Tax             &                                  0.1409 &                                  0.4597 &                                      0.1717 &                                  0.4972 &                                 0.0975 &                                        0.0789 &                   \textcolor{red}{\textbf{0.5824}} &                                       0.3336 &    \textcolor{orange}{\textbf{0.8163}} &    \textcolor{orange}{\textbf{0.8695}} &  \textcolor{PineGreen}{\textbf{0.232}} &     \textcolor{orange}{\textbf{0.1158}} \\ \hline

\end{tabular}
\caption{Mean DL distance of the 5-fold cross-validation}
\end{table*}

\begin{table*}[ht]
  \centering
  \begin{tabular}{c|l|l|l|l|l}
    Approach & Camargo (random) & Evermann & Camargo (argmax) & Tax & Francescomarino \\ \hline
    Ranking & 1.408 & 2.160 & 2.616 & 3.858 & 4.958 \\ \hline
    Probability & 36.70\% & 26.21\% & 21.43\% & 11.42\% & 4.24\% \\ \hline
  \end{tabular}
  \caption{Plackett-Luce probability of winning and ranking for the suffix prediction problem.}~\label{tab:plackett-dl}
\end{table*}

\begin{figure}[ht]
  \centering
  \includegraphics[width=0.5\textwidth]{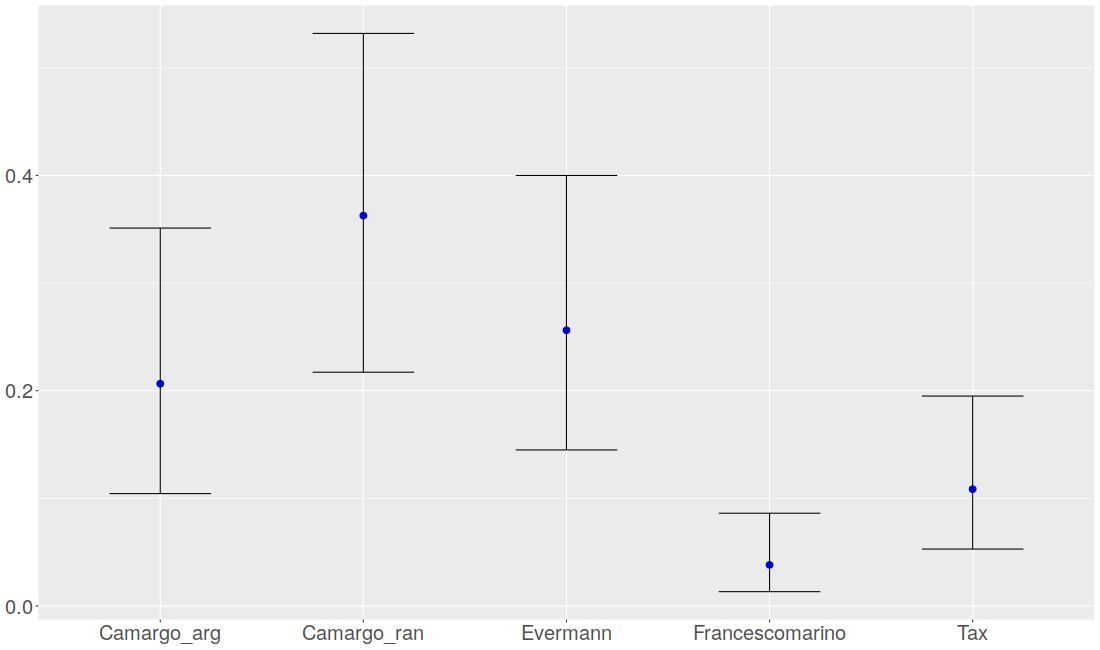}
  \caption{Credible intervals (5\% and 95\% quantiles) and expected probability of winning for the suffix prediction problem.}~\label{fig:ci-dl}
\end{figure}

\begin{figure*}[ht]
  \centering
  \subfloat[Camargo (argmax) vs Camargo (random)]{\includegraphics[width=0.3\textwidth]{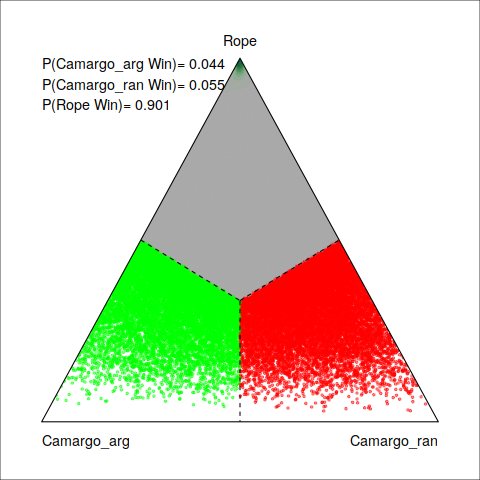}\label{fig:hie-a-rt}}
  \qquad
  \subfloat[Camargo (argmax) vs Navarin]{\includegraphics[width=0.3\textwidth]{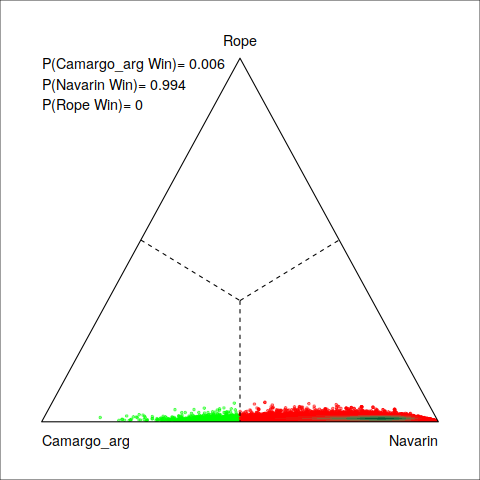}\label{fig:hie-b-rt}}
  \qquad
  \subfloat[Camargo (random) vs Navarin]{\includegraphics[width=0.3\textwidth]{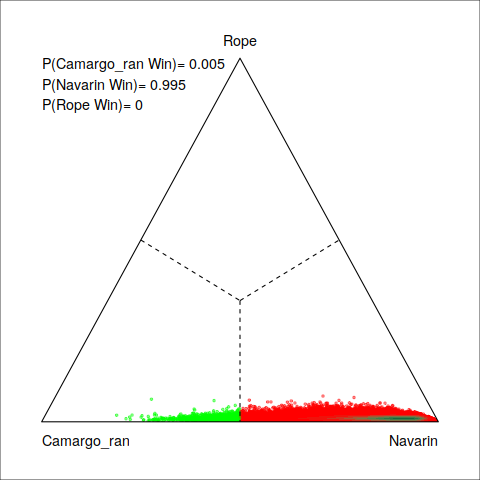}\label{fig:hie-c-rt}}
  \caption{Hierarchical Bayesian tests for the remaining time prediction problem.}~\label{fig:hierarchical-dl}
\end{figure*}

\begin{figure}[ht]
  \centering
  \includegraphics[width=0.5\textwidth]{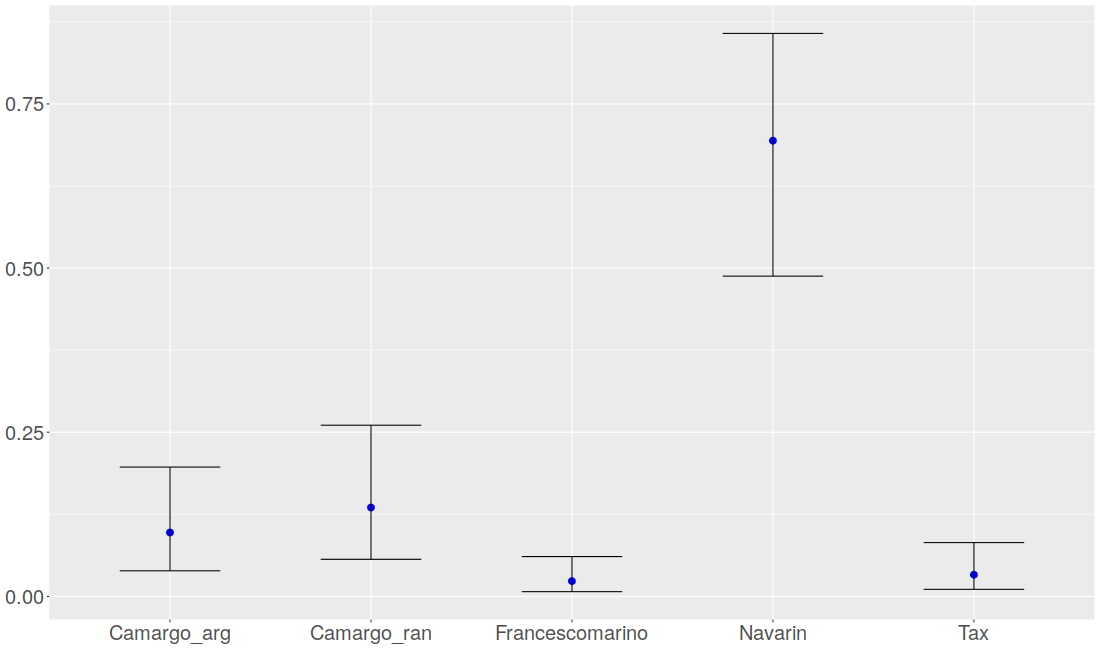}
  \caption{Credible intervals (5\% and 95\% quantiles) and expected probability of winning for the remaining time prediction problem.}~\label{fig:ci-rt}
\end{figure}

As far as the remaining time prediction problem is concerned, the best approach according to the Plackett-Luce rankings (\tablename~\ref{tab:plackett-rt}) is Navarin~\cite{Navarin2017}. This approach relies on performing a direct prediction of the remaining time for a given prefix, instead of predicting the next timestamps for a predicted suffix. Thus, the prediction errors do not compound, achieving a more precise prediction and allowing the approach from Navarin~\cite{Navarin2017} to significantly outperform the other ones, as shown in \figurename~\ref{fig:ci-rt} and as shown in the hierarchical tests (\figurename~\ref{fig:hie-b-rt} and \ref{fig:hie-c-rt}). Furthermore, in this case, there is almost no difference between the sampling procedure (``argmax'' or ``random'') used, as shown in the \figurename~\ref{fig:hie-a-rt}.

\begin{table*}[ht]
\centering
\begin{tabular}{llllllllllll}

{} &     \rotatebox{90}{BPI 2012} &   \rotatebox{90}{BPI 2012 A} & \rotatebox{90}{\shortstack[l]{BPI 2012 \\ Complete}} &   \rotatebox{90}{BPI 2012 O} &   \rotatebox{90}{BPI 2012 W} & \rotatebox{90}{\shortstack[l]{BPI 2012 \\ W Complete}} & \rotatebox{90}{\shortstack[l]{BPI 2013 \\ Closed Problems}} & \rotatebox{90}{\shortstack[l]{BPI 2013 \\ Incidents}} &             \rotatebox{90}{Env Permit} &               \rotatebox{90}{Helpdesk} &                  \rotatebox{90}{Sepsis} \\ \hline

Camargo (argmax)  &       \textcolor{red}{\textbf{29.905}} &        \textcolor{red}{\textbf{8.989}} &            \textcolor{red}{\textbf{28.045}} &        \textcolor{red}{\textbf{15.25}} &    \textcolor{orange}{\textbf{30.484}} &              \textcolor{red}{\textbf{15.263}} &                                            257.086 &          \textcolor{orange}{\textbf{28.132}} &                                   8.29 &       \textcolor{red}{\textbf{18.634}} &                                     - \\ \hline
Camargo (random)  &    \textcolor{orange}{\textbf{29.599}} &     \textcolor{orange}{\textbf{8.976}} &         \textcolor{orange}{\textbf{10.468}} &    \textcolor{orange}{\textbf{15.247}} &       \textcolor{red}{\textbf{30.564}} &           \textcolor{orange}{\textbf{10.417}} &                                            257.697 &             \textcolor{red}{\textbf{28.511}} &                                  8.277 &     \textcolor{orange}{\textbf{18.46}} &                                     - \\ \hline
Francescomarino &                                500.303 &                                 29.363 &                                     102.998 &                                 66.454 &                                 248.37 &                                         59.68 &                    \textcolor{red}{\textbf{191.1}} &                                       35.405 &        \textcolor{red}{\textbf{5.613}} &                                159.891 &       \textcolor{red}{\textbf{503.867}} \\ \hline
Navarin         &  \textcolor{PineGreen}{\textbf{7.364}} &  \textcolor{PineGreen}{\textbf{7.774}} &         \textcolor{PineGreen}{\textbf{7.2}} &  \textcolor{PineGreen}{\textbf{7.586}} &  \textcolor{PineGreen}{\textbf{7.845}} &         \textcolor{PineGreen}{\textbf{8.189}} &            \textcolor{PineGreen}{\textbf{159.164}} &       \textcolor{PineGreen}{\textbf{12.366}} &     \textcolor{orange}{\textbf{4.124}} &  \textcolor{PineGreen}{\textbf{8.186}} &  \textcolor{PineGreen}{\textbf{33.973}} \\ \hline
Tax             &                                383.054 &                                 20.283 &                                     195.482 &                                 68.155 &                                157.518 &                                       127.397 &               \textcolor{orange}{\textbf{172.849}} &                                       30.082 &  \textcolor{PineGreen}{\textbf{4.109}} &                                 68.864 &      \textcolor{orange}{\textbf{421.6}} \\ \hline

\end{tabular}
\caption{Mean remaining time of the 5-fold cross-validation}
\end{table*}

\begin{table*}[ht]
  \centering
  \begin{tabular}{c|l|l|l|l|l}
    Approach & Navarin & Camargo (random) & Camargo (argmax) & Tax & Francescomarino \\ \hline
    Ranking & 1.003 & 2.268 & 2.782 & 4.225 & 4.722 \\ \hline
    Probability & 68.60\% & 14.39\% & 10.49\% & 3.79\% & 2.72\% \\ \hline
  \end{tabular}
  \caption{Plackett-Luce probability of winning and ranking for the suffix prediction problem.}~\label{tab:plackett-rt}
\end{table*}

\section{Conclusions}
%
%

In this paper, we presented a systematic literature review of Deep Learning approaches to predictive business process monitoring. We made an analysis and classification of these approaches, supported by an exhaustive experimental evaluation of 10 approaches on 12 different process logs, according to five different perspectives: \textit{(i)} input data, \textit{(ii)} predictions, \textit{(iii)} neural network type, \textit{(iv)} sequence encoding, and \textit{(v)} event encoding.

Regarding to the input data, as shown by the experimentation, overall the usage of attributes may not benefit the predictive task at hand, except in some specific cases. We suggest to incrementally add the atributes to the predictive model and observe their effect, instead of mindlessly add them all, since their usage may hinder the convergence of the neural network. As far as the predictions is concerned, we showed how the different sampling strategies affect the predictive performance of the activity suffix prediction task, concluding that \textit{random sampling} approaches overall outperform \textit{argmax sampling} approaches for very long traces and viceversa. Regarding the remaining time problem, we recommend adopting an approach of direct prediction, since it will always outperform the approach of accumulating the predictions, due to the compounding of the prediction errors. Relating to the neural network type, we found that, overall, RNNs seem to outperform CNNs. This is due to the CNNs also often need to set up pooling layers to improve their performance but doing so adds additional hyperparameters and may impose an suboptimal architecture that may hinder their performance, while RNNs are just often stacked one after another giving good results (even though the approach of Mauro et al.~\cite{Mauro2019} seems to tackle this issue, but unsatisfactorily in some logs). However, CNNs perform way faster than RNNs, so they may be the only candidate in applications where speed is a priority (either by the abundance of data or by the need for a quick testing phase). Concerning the sequence and event encoding perspectives, even though the experimentation does not highlight differences on the performance of the multiple possibilities of encoding, there are differences on the efficiency between the encodings. For example, the \textit{one-hot} event encoding uses more memory than the \textit{embedding} event encoding when the number of different activities is very high or the \textit{continuous} sequence encoding has a faster training procedure than the \textit{prefixes padded} sequence encoding.

 We expect the experiments serve as a baseline for future works in predictive monitoring. As a future line of work, this experimentation could be enhanced with non-deep learning approaches or by integrating it with a graphical interface that eases the benchmarking procedure and visualization.

\section{Acknowledgments}
This research was funded by the Spanish Ministry for Science, Innovation and Universities (grant TIN2017-84796-C2-1-R and TIN2015-73566-JIN) and the Galician Ministry of Education, University and Professional Training (grants ED431C 2018/29 and ``accreditation 2016-2019, ED431G/08''). All grants were co-funded by the European Regional Development fund (ERDF/FEDER program). E. Rama-Maneiro is supported by the Spanish Ministry of Education, under the FPU national plan (FPU18/05687).



\bibliographystyle{IEEEtran}
%
\bibliography{survey}

%

\section{Biographies}

\vspace*{-2\baselineskip}

\begin{IEEEbiography}[{\includegraphics[width=1in,height=1.25in,clip,keepaspectratio]{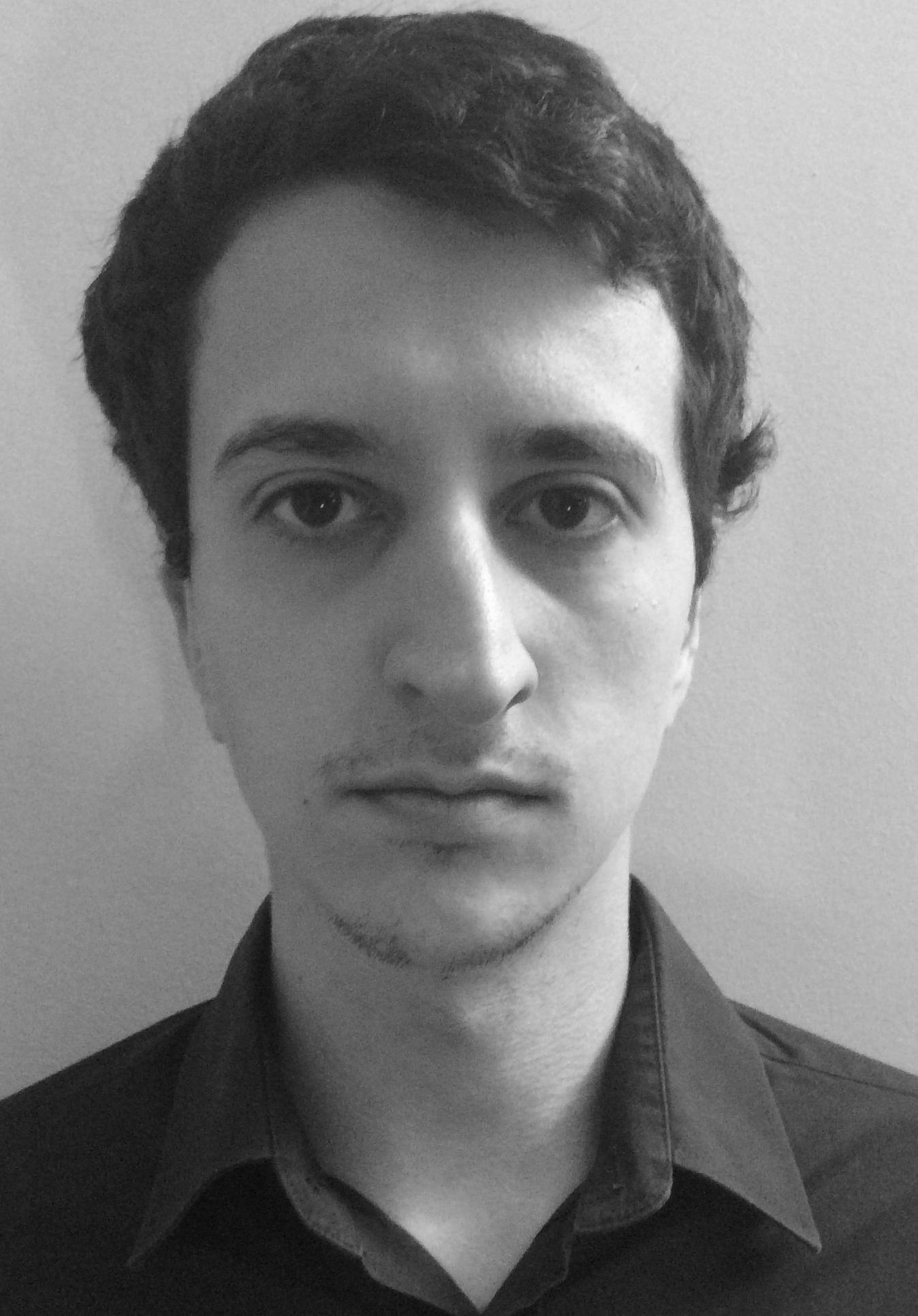}}]{EFRÉN RAMA-MANEIRO}
  received the B.Eng. degree in computer engineering and the M.Sc. degree in Big Data from the University of Santiago de Compostela, Spain in 2018 and 2019 respectively. He is a Researcher and currently working toward the Ph.D. degree at the Centro Singular de Investigación en Tecnoloxías Intelixentes, University of Santiago de Compostela. His research interests include process mining and deep learning.
\end{IEEEbiography}

\vspace*{-2\baselineskip}

\begin{IEEEbiography}[{\includegraphics[width=1in,height=1.25in,clip,keepaspectratio]{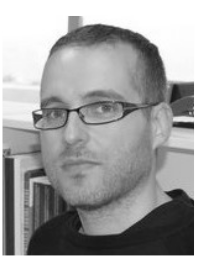}}]{JUAN C. VIDAL}
received the B.Eng. degree in computer science from the University of La Coruña, La Coruña, Spain, in 2000, and the Ph.D. degree in artificial intelligence from the University of Santiago de Compostela (USC), Santiago de Compostela, Spain, in 2010, where he was an Assistant Professor with the Department of Electronics and Computer Science, from 2010 to 2017. He is currently an Associate Researcher with the Centro Singular de Investigación en Tecnoloxías Intelixentes (CiTIUS), USC. His research interests include process mining, fuzzy logic, machine learning, and linguistic summarization.
\end{IEEEbiography}

\vspace*{-2\baselineskip}

\begin{IEEEbiography}[{\includegraphics[width=1in,height=1.25in,clip,keepaspectratio]{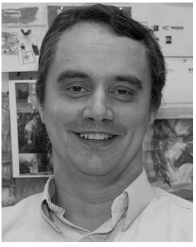}}]{MANUEL LAMA}
received the Ph.D. degree in physics from the University of Santiago de Compostela, in 2000, where he is currently an Associate Professor of artificial intelligence. He has collaborated on more than 30 projects and research contracts financed by public calls, participating as a principal investigator in 20 of them. These activities were implemented in areas, such as process discovery, predictive monitoring, and management of dynamic processes. As a result of this research, he has published over 150 scientific articles with review process in conference and national and international journals.
\end{IEEEbiography}




\end{document}